\documentclass[a4paper,fleqn, review]{cas-dc}
\usepackage{amsmath}
\usepackage[switch]{lineno}
\usepackage{graphicx}
\usepackage{subfig} 
\usepackage{tabu}  
\usepackage{hyperref}
\usepackage{multirow}
\usepackage{dsfont}
\usepackage{stfloats}
\usepackage{cite}
\usepackage{booktabs}
\usepackage{algorithm}
\usepackage{algorithmic}
\usepackage{amssymb}
\usepackage{makecell}
\usepackage{caption}
\usepackage{natbib}
\usepackage{makecell}
\usepackage{pifont}
\usepackage{bm}
\def\tsc#1{\csdef{#1}{\textsc{\lowercase{#1}}\xspace}}
\tsc{WGM}
\tsc{QE}
\tsc{EP}
\tsc{PMS}
\tsc{BEC}
\tsc{DE}
\usepackage{enumitem}

\begin{document}
\let\WriteBookmarks\relax
\def\floatpagepagefraction{1}
\def\textpagefraction{.001}
\shorttitle{}

\shortauthors{Yimin Wei et~al.}

\title [mode = title]{\textit{SARLANG-1M}: A Benchmark for Vision-Language Modeling in\\SAR Image Understanding}                      



%
\author[1,2]{Yimin Wei}[orcid=0000-0003-0100-4786]

\author[2]{Aoran Xiao}





\author[3]{Yexian Ren}

\author[4]{Yuting Zhu}

\author[1,2]{Hongruixuan Chen}

\author[2]{Junshi Xia}

\author[1,2]{Naoto Yokoya}
\cormark[1]

\affiliation[1]{organization={Graduate School of Frontier Sciences, The University of Tokyo},
    city={Chiba},
    postcode={277-8561}, 
    country={Japan}}



\affiliation[2]{organization={RIKEN Center for Advanced Intelligence Project (AIP), RIKEN},
    city={Tokyo},
    postcode={103-0027}, 
    country={Japan}}

\affiliation[3]{organization={School of Electronics and Information Engineering, Nanjing University of Information Science and Technology},
    city={Nanjing},
    postcode={430079}, 
    country={PR China}}

\affiliation[4]{organization={The School of Electronic and Communication Engineering, Sun Yat-sen University},
    city={Guangzhou},
    postcode={510006}, 
    country={PR China}}


\cortext[cor1]{Corresponding author}


\nonumnote{Manuscript submitted on May 23, 2023.}

\begin{abstract}
Synthetic Aperture Radar (SAR) is a crucial remote sensing technology, enabling all-weather, day-and-night observation with strong surface penetration for precise and continuous environmental monitoring and analysis. However, SAR image interpretation remains challenging due to its complex physical imaging mechanisms and significant visual disparities from human perception. Recently, Vision-Language Models (VLMs) have demonstrated remarkable success in RGB image understanding, offering powerful open-vocabulary interpretation and flexible language interaction. However, their application to SAR images is severely constrained by the absence of SAR-specific knowledge in their training distributions, leading to suboptimal performance. To address this limitation, we introduce \textit{SARLANG-1M}, a large-scale benchmark tailored for multimodal SAR image understanding, with a primary focus on integrating SAR with textual modality. \textit{SARLANG-1M} comprises more than 1 million high-quality SAR image-text pairs collected from over 59 cities worldwide. It features hierarchical resolutions (ranging from 0.1 to 25 meters), fine-grained semantic descriptions (including both concise and detailed captions), diverse remote sensing categories (1,696 object types and 16 land cover classes), and multi-task question-answering pairs spanning seven applications and 1,012 question types. Extensive experiments on mainstream VLMs demonstrate that fine-tuning with \textit{SARLANG-1M} significantly enhances their performance in SAR image interpretation, reaching performance comparable to human experts.
The dataset and code will be made publicly available at {\url{https://github.com/Jimmyxichen/SARLANG-1M}}.
\end{abstract}


\begin{keywords}
\sep Synthetic Aperture Radar (SAR) \sep 
Vision-Language Models (VLMs) \sep 
High-quality Texts \sep Image Captioning \sep Visual Question Answering (VQA) \sep

\end{keywords}

\maketitle

\section{Introduction}\label{sec:1}


\par Synthetic Aperture Radar (SAR) is a vital remote sensing technology renowned for its ability to capture high-resolution images under all weather conditions, including clouds, rain, and darkness. Unlike optical imagery, SAR utilizes microwave signals that can penetrate materials such as vegetation, dry soil, and man-made structures, depending on the wavelength. Longer wavelengths, like L-band or P-band, enable SAR to penetrate forest canopies and reveal subsurface features, making it invaluable for forestry monitoring, archaeological exploration, and soil moisture analysis \citep{gao2017synergetic,lucas2012global}. Combined with its all-weather and day-night imaging capabilities, SAR's penetration ability makes it an essential tool for applications such as disaster management, environmental monitoring, and military surveillance \citep{reamer1993new,sharma2008sar}. These unique advantages have spurred growing research interest in advancing robust SAR image interpretation. 



\par Despite significant advancements in various SAR-related tasks such as object detection \citep{wang2019sar, chai2024enhanced}, semantic segmentation \citep{pena2024deepaqua, wu2024ctmanet}, and other areas \citep{zhao2024sea}, SAR image interpretation remains highly challenging. One of the primary difficulties arises from the fundamental differences between SAR and optical imagery \citep{richards2009remote,woodhouse2017introduction}. Unlike RGB images, which capture visible light, SAR images are generated through microwave backscattering, where the radar system emits microwave pulses toward the Earth's surface and captures the reflected signals. This process introduces distinctive characteristics such as speckle noise and geometric distortions that complicate image analysis. Additionally, the inherent semantic complexity of SAR images—often characterized by intricate textures and non-intuitive patterns—makes interpretation more challenging than that of natural optical imagery. Moreover, the disparity between human visual perception and SAR representations further complicates annotation, requiring domain expertise and extensive training. Due to the high costs \citep{ball2017comprehensive} and specialized knowledge required for annotation, existing SAR datasets remain limited in both size and diversity. These challenges underscore the need for specialized datasets, algorithms, and models tailored to SAR data, driving ongoing research and innovation in the field.

\par
\begin{figure}[!t]
\centering
\includegraphics[width=3.3in]{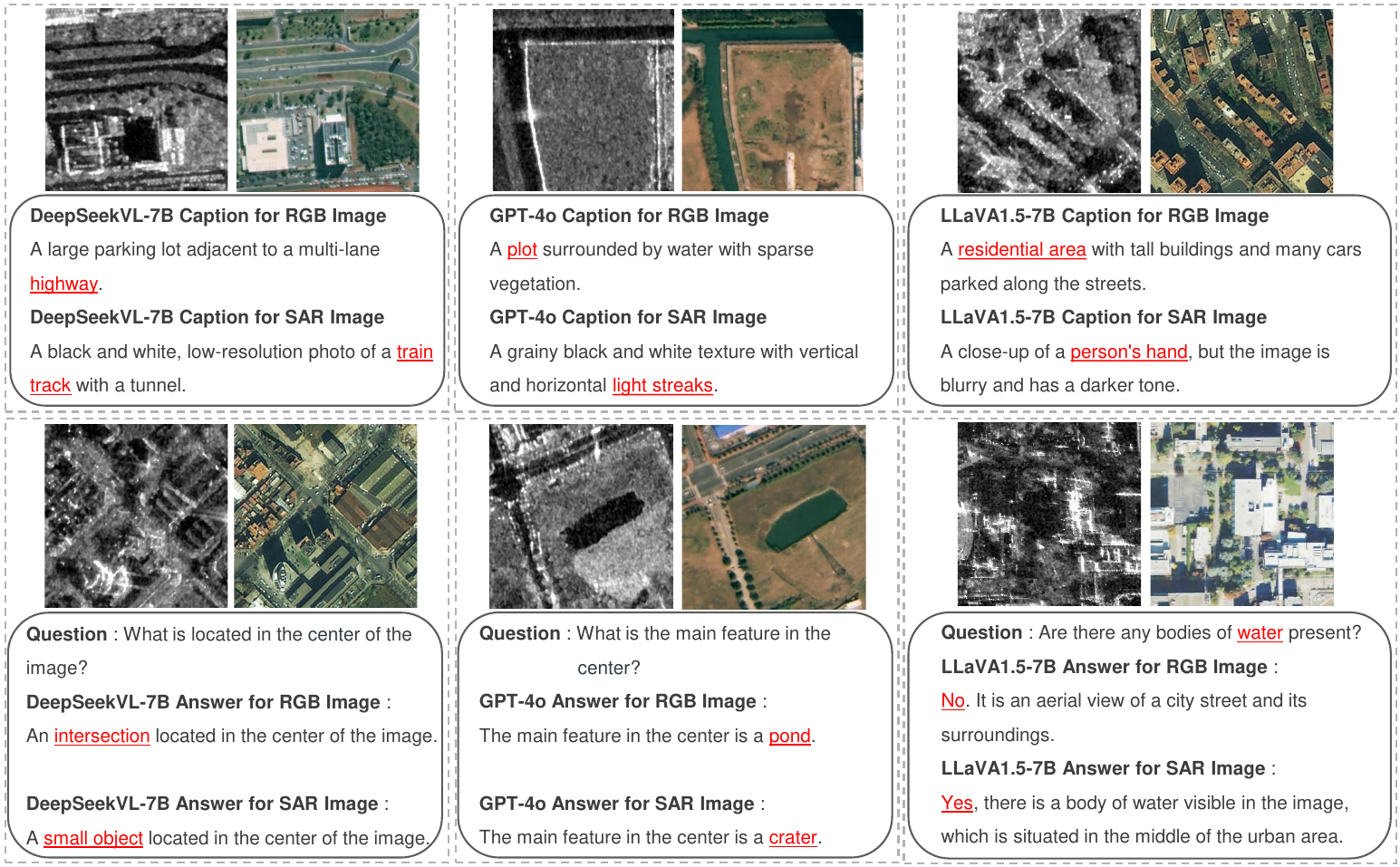}
\caption{\textbf{Examplar Cases of Mainstream VLMs that Exhibit Strong Performance on RGB images but Struggle with Paired SAR Images.} The first row demonstrates the image captioning results generated by the DeepSeekVL-7B \citep{lu2024deepseek}, GPT-4o \citep{achiam2023gpt}, and LLaVA1.5-7B \citep{liu2024visual} models, with the left, center, and right examples illustrating their respective outputs. The second row presents the VQA results produced by the same models, with the left, center, and right examples further highlighting the challenges associated with SAR image interpretation. Key terms in the text are highlighted with red underlines for emphasis. For instance, the LLaVA1.5-7B \citep{liu2024visual} model, as shown in the right example in the first row, misclassifies a residential scene as a person’s hand. Furthermore, in the right example in the second row, the LLaVA1.5-7B \citep{liu2024visual} model fails to identify the presence of water in the SAR image, despite successfully recognizing it in the corresponding RGB image.}
\label{fig:fig0}
\end{figure}

 Recently, VLMs \citep{hu2023rsgpt, kuckreja2023geochat} have demonstrated significant potential in enhancing optical RGB image understanding through large-scale pre-training and instruction tuning. Unlike traditional models, which are often restricted to fixed vocabularies and predefined labels, VLMs exhibit superior image interpretation capabilities due to several key advantages. First, they enable flexible language interaction, allowing for more intuitive and context-aware image comprehension. Second, through multimodal pre-training on large-scale datasets, VLMs acquire extensive open-vocabulary language capabilities, enabling them to describe a wide range of scenes and concepts beyond predefined categories. The rapid advancement of VLMs has created new opportunities for SAR image interpretation. However, directly applying existing VLMs to SAR image analysis is impractical. These models are predominantly trained on RGB image datasets, meaning SAR images fall outside the distribution of their training data. As a result, VLMs perform poorly when applied to SAR images, which exhibit fundamentally different data characteristics from RGB imagery. An example of this performance gap is illustrated in Figure \ref{fig:fig0}.



\par To enable VLMs to effectively interpret SAR images, the development of a large-scale SAR-text dataset is both essential and urgent. However, constructing such a dataset presents significant challenges due to data scarcity and the complexity of annotation. One of the primary obstacles is the lack of large-scale SAR image datasets with high-quality text annotations. Annotating SAR images requires expert-level supervision, as the unique characteristics of SAR images demand specialized domain knowledge for accurate interpretation. The high cost and complexity associated with SAR annotation have severely constrained the scale and diversity of existing datasets, thereby limiting the advancement of SAR-specific VLMs.

\par
\begin{figure*}[!t]
\centering
\includegraphics[width=6.8in]{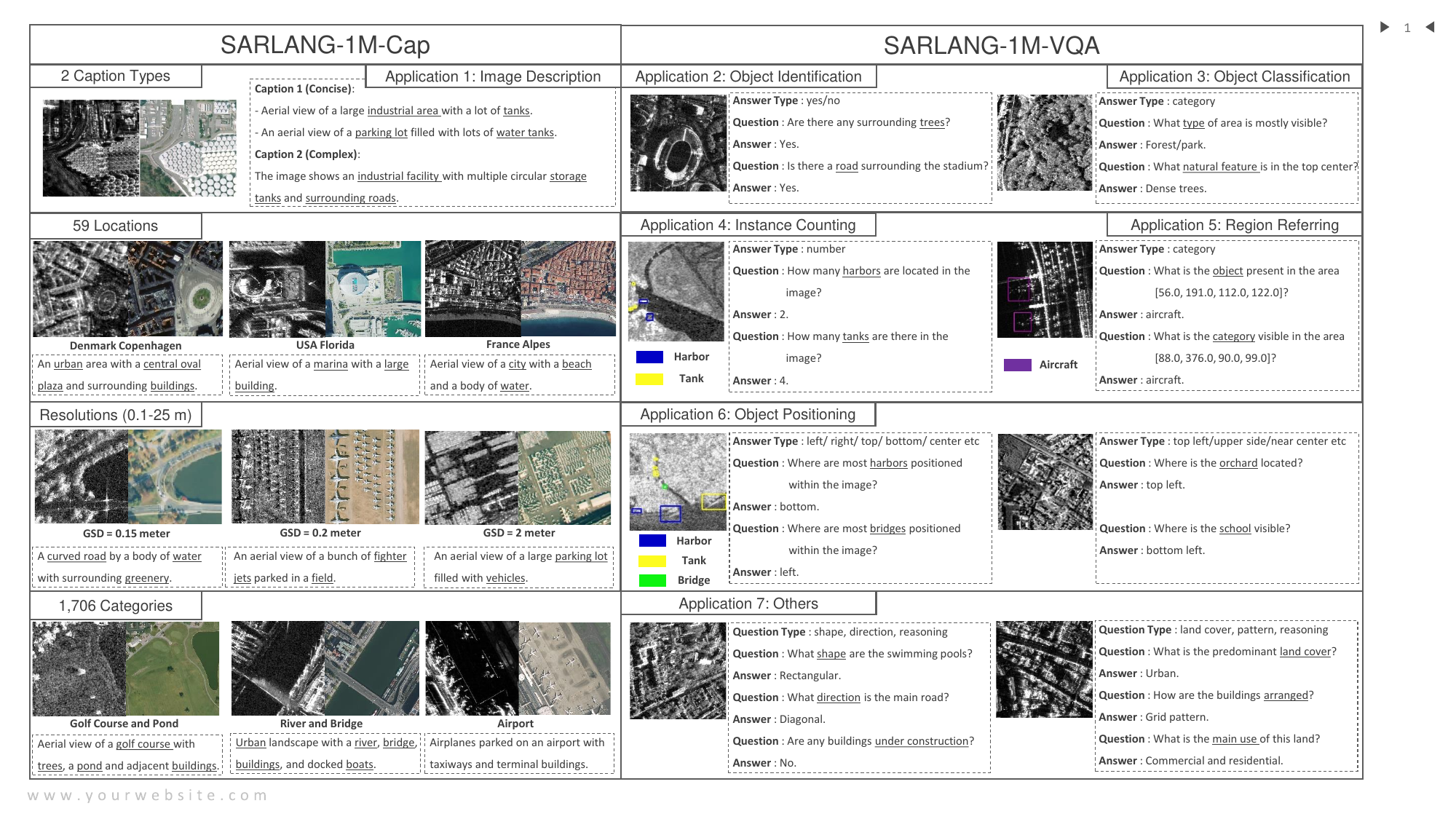}
\caption{
\textbf{Examples from the \textit{SARLANG-1M} dataset.} \textit{SARLANG-1M} consists of two benchmarks: \textit{SARLANG-1M-Cap} for SAR image captioning and \textit{SARLANG-1M-VQA} for SAR image VQA. The \textit{SARLANG-1M-Cap} benchmark is characterized by four dimensions: caption type, presence of location information, resolution, and number of categories, supporting the task of detailed SAR image description. In contrast, the \textit{SARLANG-1M-VQA} benchmark is designed to enable six additional applications, including object identification, object classification, instance counting, region referring, object positioning and others.
}  
\label{fig:fig12}
\end{figure*}

 To overcome these challenges and advance SAR image understanding, this paper introduces \textit{SARLANG-1M}, a novel large-scale benchmark designed for multimodal SAR image interpretation, as illustrated in Figures \ref{fig:fig12}. The \textit{SARLANG-1M} dataset consists of approximately 1 million high-quality SAR image-text pairs collected from over 59 cities worldwide. It encompasses multi-scale resolutions ranging from 0.1 to 25 meters, fine-grained semantic descriptions (including both concise and detailed captions), a diverse set of remote sensing categories (1,696 object categories and 16 land cover categories), and multi-task question-answering pairs covering seven applications and 1,012 question types.


\par To generate high-quality and domain-specific text descriptions, we constructed the \textit{SARLANG-1M} dataset using two distinct text generation approaches. The first approach involves text transfer through modality pairs. Specifically, we collected a large set of paired optical RGB and SAR images, where state-of-the-art VLMs (e.g., GPT-4o \citep{achiam2023gpt}) were used to generate high-quality text annotations for the optical RGB images. These annotations were then naturally aligned with their corresponding SAR images. To ensure accuracy, we manually reviewed and filtered out incorrect or irrelevant descriptions, such as references to color information that are applicable to RGB images but not to SAR images. This approach enables the transfer of knowledge from the RGB domain to SAR modalities, producing comprehensive and precise text annotations. However, existing VLMs exhibit limited spatial reasoning capabilities, particularly in identifying positional and quantitative attributes of objects—key aspects for remote sensing interpretation. To address this limitation, our second strategy involves generating fine-grained text annotations from existing bounding box annotations in SAR object detection datasets \citep{li2024sardet}. By combining this approach with rigorous manual verification by SAR experts, the \textit{SARLANG-1M} dataset comprises 118,331 SAR images with 1,126,277 human-verified text annotations. This dataset facilitates the training and evaluation of VLMs across various SAR image understanding tasks, including VQA and image captioning.

\par Furthermore, we conduct comprehensive experiments using the \textit{SARLANG-1M} dataset, evaluating both traditional models and ten state-of-the-art VLMs(e.g., DeepSe\-ek-VL \citep{lu2024deepseek} and QWEN2.5-VL \citep{Qwen2.5-VL}), demonstrating the dataset's effectiveness in enhancing VLMs' ability to interpret SAR images. Specifically, \textit{SARLANG-1M} achieves significant performance improvements, with increases of 67.20\% in CIDEr \citep{vedantam2015cider} and 40.22\% in GPT-4-based accuracy \citep{li2024vrsbench} for SAR image captioning and SAR VQA tasks, respectively. The visualization results of VLM’s predictions across seven SAR image applications further validate the effectiveness of our \textit{SARLANG-1M} dataset. Additionally, per-word Grad-CAM \citep{selvaraju2017grad} visualizations are provided to demonstrate that \textit{SARLANG-1M} enables VLMs to better align textual keywords with the visual content in SAR images.

\par The key contributions of our work are summarized as follows:



\par 1) \textbf{Comprehensive Large-Scale Dataset:} \textit{SARLANG-1M} consists of 118,331 multi-resolution SAR images, providing 1,080,627 high-quality image-text pairs and 45,650 detailed captions from over 59 cities worldwide. The dataset spans multi-scale resolutions from 0.1 to 25 meters and includes fine-grained semantic descriptions (both concise and detailed captions), diverse remote sensing categories (1,696 object categories and 16 land cover types), and multi-task question-answering pairs covering 1,012 question types.


\par 2) \textbf{Benchmark for Vision-Language Tasks:} The \textit{SAR\-LANG-1M} benchmark comprises two key components: \textit{SAR\-LANG-1M-Cap} for image captioning task and \textit{SARLA\-NG-1M-VQA} for VQA task. As the largest SAR image-text dataset to date, it supports seven critical remote sensing applications, including image description, object identification, object classification, instance counting, region referring, and object positioning.


\par 3)  \textbf{Extensive Model Evaluation and Improvement:} We conduct a comprehensive performance evaluation using \textit{SARLANG-1M}, assessing two state-of-the-art traditional models and ten VLMs. Experimental results demonstrate that fine-tuning mainstream VLMs on \textit{SARLANG-1M} significantly enhances their performance on SAR-specific vision-language tasks, achieving results comparable to those of human experts.



\par The remainder of this paper is organized as follows. Section \ref{sec:related_work} reviews related work. Section \ref{sec:Dataset} details the construction progress of the \textit{SARLANG-1M} dataset. The experimental results and analysis are provided in Section \ref{sec:experiments}. Finally, Section \ref{sec:conclusion} draws the conclusion.


\section{Related Works}
\label{sec:related_work}
\subsection{SAR for Perception}

\par Early SAR image processing models typically relied on manually crafted features and simple statistical learning models to categorize these features into predefined classes or patterns \citep{du2015random,zabalza2014novel}. However, these approaches were heavily dependent on domain expertise for feature creation, rendering it less effective for complex tasks. As a result, the scalability and generalizability of these models were significantly constrained. The advent of deep learning, characterized by end-to-end trainable neural networks, has transformed SAR image analysis by eliminating the need for intricate feature extraction processes. This advancement has led to higher accuracy and enhanced model robustness. Prominent deep learning architectures such as Convolutional Neural Networks (CNNs) \citep{szegedy2015going}, Long Short-Term Memory networks (LSTMs) \citep{graves2012long},Transformers \citep{vaswani2017attention}, Diffusion models \citep{ho2020denoising}, and the You Only Look Once (YOLO) \citep{redmon2016you} framework have significantly advanced the field of SAR image understanding. These models have improved performance across various tasks, including classification \citep{zhang2024lightweight,liu2024softformer,geng2024polarimetric,zhang2024multiscale}, semantic segmentation \citep{ristea2024multi,zhou2025transformer,liu2024review}, object detection \citep{wang2024detection,tang2024dbw,chen2024gcn,chen2024yolo}, and image-to-image translation \citep{zhao2024hvt,rangzan2024sar,zhang2024sar,guo2024ms}.

\subsection{Text Generation for Remote Sensing}

\subsubsection{Image Captioning}

\par Remote sensing image captioning (RSIC) aims to describe key objects within remote sensing images using natural language. RSIC faces several challenges, including scale variations and cross-modal disparities. Objects belonging to the same category may appear in different scales within remote sensing images, and significant modality differences exist between the input image and the output text. This necessitates precise alignment and effective fusion of image and text features during text generation.

\par Current RSIC models can be categorized into three primary approaches: retrieval-based methods, syntax-template-based methods, and encoder-decoder-based methods. Retrie\-val-based methods attempt to generate captions by retrieving similar images and using their annotated sentences \citep{wang2020retrieval, wang2019semantic}. Syntax-template-based methods involve detecting objects in images, identifying candidate words, and filling them into predefined syntax templates \citep{shi2017can}. However, the sentences generated by these two methods often lack flexibility and naturalness.

\par Recently, encoder-decoder-based approaches have been developed to learn representations of remote sensing images and grammar automatically, allowing for the creation of more adaptable sentences. These models generally operate in two stages: vision feature extraction and text generation. In the vision feature extraction stage, CNNs \citep{szegedy2015going} or Vision Transformers \citep{liu2021swin} are typically employed to extract semantic features from images and encode them into high-dimensional vectors. For text generation, Recurrent Neural Networks (RNNs) \citep{ rumelhart1986learning} or Transformers \citep{vaswani2017attention} are used to transform these feature vectors into text.
Efforts to enhance feature extraction include the work by Zheng et al. \citep{zheng2023farseg++}, who introduced attention-based feature fusion modules alongside an auxiliary object detection task to enrich visual feature representation. Similarly, Ye et al. \citep{ye2022joint} proposed a multilabel classification task to capture prior knowledge, complemented by a semantic gate module to guide the generation of hidden states.

\par In terms of enhancing the decoder, Hoxha and Melgani \citep{hoxha2021novel} developed an SVM-based decoder capable of transforming visual vectors into high-quality textual descriptions, especially for small datasets. Furthermore, Transformer-based techniques have been explored: Chen et al. \citep{chen2022typeformer} constructed a Swin Transformer network to explore image features, using the Transformer network as a decoder to generate captions. Liu et al. \citep{liu2022remote} proposed a multilayer aggregated transformer as the decoder, with LSTM networks linking the transformer encoder and decoder for improved feature representation. Yang et al. \citep{yang2024hcnet} utilized a dual-LSTM architecture as the decoder for generating coherent and contextually relevant captions. Instead of merely concatenating generated visual features, word features, and LSTM hidden states for input into the LSTM network, a cross-modal feature alignment module (CFMI) is used to align visual features with other input features before fusion. Additionally, a cross-modal feature alignment loss is introduced to minimize the distance between encoded visual features and LSTM-encoded sentence features.

\subsubsection{VQA}

\par Visual Question Answering for Remote Sensing (RSVQ\-A) focus on generating accurate answers to natural language inquiries by comprehensively analyzing the visual content of remote sensing images. The standard framework for addressing the RSVQA problem entails the integration of an aerial image with a corresponding natural language question to output a contextually relevant answer derived from the image's content. The types of questions typically encountered in RSVQA include, but are not limited to, inquiries regarding categories, quantities, characteristics, spatial positions, and purposes of key objects within the aerial scene, as well as the relative relationships between distinct objects. 
The research of RSVQA was initiated by Sylvain et al. \citep{lobry2020rsvqa}, who introduced a novel RGB-text dataset derived from Sentinel-2 images, alongside a deep learning framework to assess the dataset's efficacy. This framework is systematically organized into three distinct stages: feature extraction, feature fusion, and prediction based on the fused features. During the feature extraction phase, a CNN-based model is utilized to extract visual features from the input remote sensing images, while a RNN-based model is employed to embed a sentence of words into a latent space, thereby generating a linguistic embedding. Specifically, ResNet-152 \citep{he2016deep}, pretrained on ImageNet \citep{deng2009imagenet}, serves as the visual encoder, while the skip-thoughts model \citep{kiros2015skip}, trained on the BookCorpus dataset \citep{zhu2015aligning}, functions as the language encoder. The extracted visual feature and textural embedding are subsequently fused via point-wise multiplication in the feature fusion stage, followed by a projection onto a fixed-dimensional vector using a Multi-Layer Perceptron (MLP) layer.
The follow-up works focus on enhancing this framework by substituting the convolutional and recurrent encoders with more advanced architectures. For instance, Silva et al. \citep{silva2022remote} integrated the Multimodal Medical BERT (MMBERT) \citep{khare2021mmbert} with an EfficientNetV2 \citep{tan2021efficientnetv2} image encoder and a RealFormer \citep{he2020realformer} multimodal encoder. Similarly, Tosato et al. \citep{tosato2024segmentation}  embed an attention mechanism \citep{vaswani2017attention} guided by segmentation into the RSVQA pipeline, positing that segmentation critically guides attention by delivering a contextual comprehension of the visual information, thereby emphasizing specific objects or areas of interest. In their work, the ResNet-50 \citep{he2016deep} and DistilBERT \citep{sanh2019distilbert} models are leveraged as the visual and textual encoders, respectively.

\subsection{Vision-Language Models (VLMs)}

\subsubsection{General VLM}
\par Foundation models have recently emerged as a central topic in AI research due to their ability to address a variety of downstream tasks after undergoing extensive pretraining. A Vision Foundation Model (VFMs) specifically targets visual tasks, while a VLMs enhances VFMs by integrating both visual and textual modalities, facilitating more complex multimodal reasoning and a wider range of applications. The rise of expansive large-scale image-text datasets has propelled VLMs into the spotlight, as they have demonstrated exceptional performance on multimodal tasks such as image captioning, VQA, and cross-modal retrieval. Typically, mainstream VLMs comprise a pre-trained visual encoder to process visual data, a language encoder to interpret user instructions and generate responses, and a vision-language cross-modal connector to fuse visual feature representations with textual embeddings. A key component in pre-training VLMs is the effective bridging of vision and language through image-text pairs. This is generally accomplished by employing two primary objectives: contrastive learning and generative modeling.

\par Contrastive Learning, exemplified by the CLIP \citep{radford2021learning} model, is designed to align image-text pairs by maximizing the similarity between matched pairs while minimizing it for mismatched ones. This methodology allows CLIP to develop a unified representation space for both visual and textual data, leading to outstanding zero-shot classification capabilities. Following this, OpenCLIP \citep{openCLIP2022} offers a fully open-source version of the CLIP model. Building on CLIP’s architecture, EVA-CLIP \citep{sun2023eva} introduces enhanced training strategies aimed at improving performance, reducing computational overhead, and increasing training stability.

\par Generative Modeling focuses on training models to produce coherent and relevant text or images, primarily through two methods: masked reconstruction and autoregressive next-token prediction. Masked reconstruction, employed by models like FLAVA \citep{singh2022flava} and MaskVLM \citep{kwon2022masked}, involves predicting masked tokens within text or occluded patches in images. This technique enhances the model's ability to understand context and relationships across visual and linguistic modalities, thereby improving cross-modal comprehension. Autoregressive next-token prediction, a dominant paradigm in VLMs training, aims to generate the next token in a sequence using prior context. This approach typically involves three core components: a pre-trained language module such as Llama3 \citep{dubey2024llama} or Vicuna \citep{vicuna2023}, a pre-trained visual encoder like EVA-CLIP \citep{sun2023eva} or CLIP \citep{radford2021learning}, and a trainable connection module to bridge visual and language embeddings, such as MLP layers \citep{LLaVAnext2024} or Q-Former \citep{li2023blip}. This architecture allows the model to leverage pre-trained visual and language capabilities while the connection module learns to align information between these modalities.  LLaVA \citep{liu2024visual} exemplifies pioneering work in this area by integrating and surpassing the strengths of both Vicuna \citep{vicuna2023} and CLIP. It achieves this by connecting CLIP's open-set visual encoder with Vicuna's language decoder, then fine-tuning end-to-end on generated instructional vision-language data.
Additionally, other approaches \citep{li2022blip,li2023blip,instructblip2023} are designed to enhance vision-language representation learning by employing multiple training objectives. For instance, BLIP  \citep{li2022blip} employs a pre-training strategy akin to CLIP but utilizes distinct training objectives. Specifically, BLIP incorporates contrastive learning to align visual and textual information by pulling together corresponding image-text pairs and pushing apart non-corresponding ones. However, BLIP also employs techniques such as image-text matching and captioning objectives to enhance its learning process, thus improving the representations derived from image-text pairs. This multi-objective strategy allows BLIP to capture rich and nuanced relationships between images and text beyond the scope of contrastive learning alone.

\subsubsection{VLM for Remote Sensing}

\par Recently, VLMs \citep{liu2024remoteclip,kuckreja2024geochat} have been applied to the field of remote sensing imagery, demonstrating significant potential in addressing various downstream tasks such as zero-shot classification, image captioning, VQA, and object referencing. Building on the CLIP \citep{radford2021learning} architecture, Liu et al. \citep{liu2024remoteclip} collected a comprehensive remote sensing dataset comprising 17 sub-datasets to pretrain the RemoteCLIP \citep{liu2024remoteclip} model, subsequently evaluating its performance across diverse downstream tasks. In contrast to the CLIP \citep{radford2021learning} model, RemoteCLIP \citep{liu2024remoteclip} employs the InfoNCE loss function to calculate the similarity between visual and textual features. Similarly, Kuckreja et al. \citep{kuckreja2024geochat} developed a remote sensing multimodal instruction-following dataset that includes images with brief descriptions generated by Vicuna-v1.5 \citep{vicuna2023}. Using this dataset, Kuckreja et al. fine-tuned the LLaVA-1.5 \citep{liu2024visual} model with LoRA \citep{hu2021lora} technology.
Recognizing the capability of MiniGPT-4 \citep{zhu2023minigpt} in training a single projection layer to effectively align visual features with large language models (LLMs), RSGPT \citep{hu2023rsgpt} was developed by fine-tuning InstructBLIP \citep{instructblip2023} on their RSICap dataset. By fine-tuning only the Q-Former \citep{li2023blip} network and the linear layer of InstructBLIP, the RSGPT model efficiently learns to align visual features of remote sensing images with LLMs in a data-efficient manner.
The advancement of VLMs for remote sensing imagery has been facilitated by the availability of large-scale RGB image-text datasets \citep{liu2024remoteclip,li2024vrsbench}. However, to the best of our knowledge, there is a scarcity of SAR image-text datasets due to the high cost and complexity of SAR annotation, which significantly limits the development of VLMs in the SAR image domain.

\begin{figure*}[!t]
\centering
\includegraphics[width=6.8in]{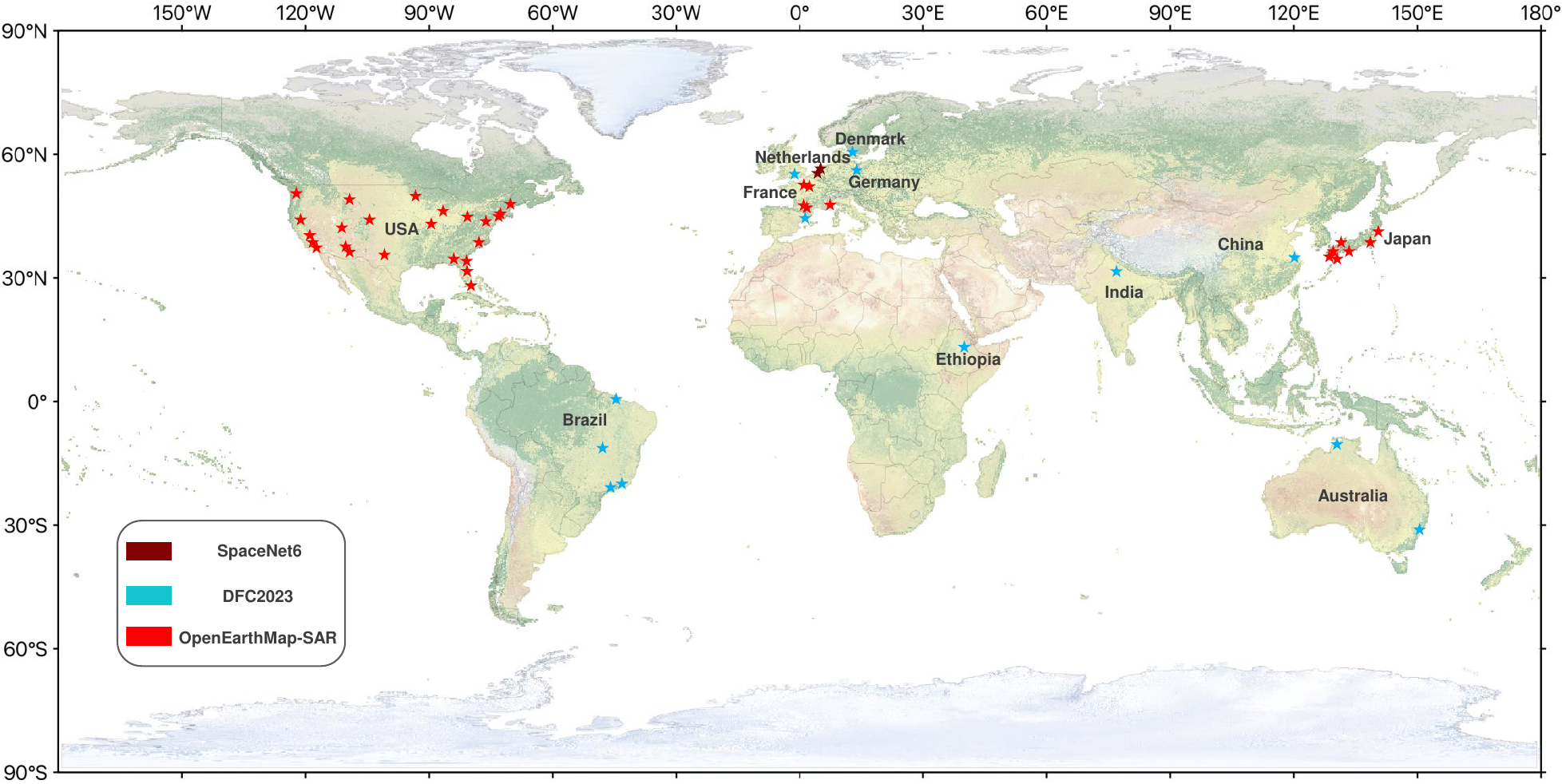}
\caption{\textbf{Cities included in the \textit{SARLANG-1M} dataset. \textit{SARLANG-1M} contains SAR images from over 59 cities worldwide, most of which are highlighted on the map.}}
\label{fig:fig34}
\end{figure*}

\section{\textit{SARLANG-1M} dataset}
\label{sec:Dataset}
\subsection{Dataset Description}


\par In this section, we provide a comprehensive overview of the \textit{SARLANG-1M} dataset, detailing its geographic coverage, sensor types, resolution scales, and category distribution. Additionally, we present a comparative analysis between \textit{SARLANG-1M} and existing SAR interpretation datasets, highlighting its unique advantages and contributions to the field.




\noindent \textbf{Dataset Overview.} \textit{SARLANG-1M} consists of 118,331 SAR images, covering more than 59 cities worldwide, ensuring diverse geographic representation, as shown in Figure \ref{fig:fig34}. The dataset integrates SAR images from four distinct sources, collected by over 12 different satellites, as summarized in Table \ref{tab:table1}. These images capture a broad range of common remote sensing environments, including airports, harbors, rivers, schools, residential areas, forests, and commercial zones, making it highly representative of real-world SAR imaging scenarios.


\par To support advancements in SAR image understanding under VLM era, \textit{SARLANG-1M} is designed for two key tasks: SAR image captioning (\textit{SARLANG-1M-Cap}) and SAR image VQA (\textit{SARLANG-1M-VQA}). The SAR image captioning benchmark provides two levels of textual descriptions: concise captions, which offer brief but informative summaries of image content, and detailed captions, which provide richer semantic descriptions with more contextual information. The \textit{SARLANG-1M-VQA} benchmark is structured into six primary question types: object identification, object classification, instance counting, region referring, object positioning, and general inquiries. These tasks are designed to challenge and evaluate the capabilities of VLMs in SAR-specific interpretation, ensuring a comprehensive assessment across various remote sensing applications.



\noindent \textbf{Comparison with Existing Datasets.} Table \ref{tab:table5} presents a comprehensive comparison between \textit{SARLANG-1M} and existing SAR datasets, highlighting its unique advantages. As shown in the table, \textit{SARLANG-1M} offers a significantly broader range of spatial resolutions and a substantially larger number of annotated samples compared to previous datasets. One of the key distinguishing features of \textit{SARLANG-1M} is the inclusion of high-quality text annotations, which enable advanced SAR image captioning and VQA tasks. This rich textual information bridges the gap between vision and language in SAR interpretation, providing a valuable resource for training and evaluating multimodal models.

\begin{table*}[t]
\renewcommand{\arraystretch}{1.35}
\centering
\newcommand{\tabincell}[2]{\begin{tabular}{@{}#1@{}}#2\end{tabular}}
\caption{\label{tab:table1}{\bf Detailed Information of Four Public Datasets used to Construct our \textit{SARLANG-1M} dataset, including Polarizations, Resolution, Collected Sensors and SAR Image Number.} '\# P' denote the polarizations.}
\scalebox{0.9}{
\begin{tabular}{|c|c|c|c|c|}
\hline

{\bf Dataset} & {\bf \# P} & {\bf Resolution (m)} & {\bf Sensor} & {\bf \# SAR Image} 

\\\hline
{SpaceNet6 \citep{shermeyer2020spacenet}} & {4} & {0.5} & {Capella Space's Satellite} & {3,311} 
\\\hline

{DFC2023 \citep{persello20232023}} & {1} & {0.5, 0.8, 1, 2} & \begin{tabular}[c]{@{}l@{}}SuperView-1 ("GaoJing" in Chinese) \\ Gaofen-2, Gaofen-3 Satellites \end{tabular} & {5,222} 

\\\hline

{OpenEarthMap-SAR \citep{xia2025openearthmap}} & {1} & {0.15 $\sim$ 0.5} & {Umbra Satellites}  & {4,813} 

\\\hline

{SARDet-100k \citep{li2024sardet}} & {1} & {0.1 $\sim$ 25} & \begin{tabular}[c]{@{}l@{}}Gaofen-3, Sentinel-1, TerraSAR-X \\ RadarSat-2, TanDEMX, HISEA-1 \\ Airborne SAR Synthetic Slice\end{tabular} & {104,985} 

\\\hline
\end{tabular}}
\end{table*}











\begin{table*}[h!]
\renewcommand{\arraystretch}{1.35}
\centering
\caption{\label{tab:table5}{\bf Comparison with Existing SAR Datasets in terms of Sample Numbers, Resolutions, Supported Tasks, and Text Availability.}}
\scalebox{0.85}{
\begin{tabular}{|l|c|c|c|l|c|}
\hline
\textbf{Dataset}     & \textbf{Proposed Time} & \textbf{\# Samples} & \textbf{Resolution (m)} & \textbf{Task} & \textbf{Provide Texts} \\ \hline
OpenSARShip \citep{huang2017opensarship}          & 2018                   & 11,346           & 10                      & Classification & \ding{55}                      \\ \hline
FUSAR-Map \citep{shi2021object}            & 2021                   & 610              & 3                       & Semantic Segmentation & \ding{55}                     \\ \hline
SSDD \citep{zhang2021sar}                 & 2021                   & 1,160            & 1 $\sim$ 15             & Ship Detection & \ding{55}                      \\ \hline
MSAR \citep{xia2022crtranssar}                 & 2022                   & 28,449           & $\leq$ 1                & Object Detection & \ding{55}                      \\ \hline
SADD \citep{zhang2022sefepnet}                 & 2022                   & 2,966            & 0.5 $\sim$ 3            & Airplane Detection & \ding{55}                      \\ \hline
SAR-AIRcraft \citep{zhirui2023sar}         & 2023                   & 4,368            & 1                       & Object Detection & \ding{55}                      \\ \hline
SIVED \citep{lin2023sived}                & 2023                   & 1,044            & 0.1, 0.3                & Object Detection & \ding{55}                      \\ \hline
SARDet-100k \citep{li2024sardet}         & 2024                   & 104,985          & 0.1 $\sim$ 25           & Object Detection & \ding{55}                      \\ \hline
BRIGHT \citep{chen2025bright} & 2025                   & 4,246            & 0.3 $\sim$ 1         & \begin{tabular}[c]{@{}l@{}}Change Detection, \\ Image Mataching, \\ SSL, UDA\end{tabular} & \ding{55} \\ \hline

OpenEarthMap-SAR \citep{xia2025openearthmap}              & 2025                   & 5,033            & 0.15 $\sim$ 0.5         & \begin{tabular}[c]{@{}l@{}}Semantic \\ Segmentation, \\ Image Translation, \\ UDA\end{tabular} & \ding{55} \\ \hline
Ours                 & 2025                   & 1,126,277          & 0.1 $\sim$ 25           & Image Captioning, VQA &  \checkmark       \\ \hline
\end{tabular}}
\end{table*}




\par To enable a fair and objective comparison, we introduce \textit{overall similarity} as a metric to assess the redundancy and diversity of \textit{SARLANG-1M} in relation to existing SAR datasets. The \textit{overall image similarity} metric quantifies dataset quality by measuring the internal similarity of images, providing insights into the dataset’s distributional uniqueness. Specifically, we compute the pairwise cosine similarity of image features for all image pairs within the dataset and aggregate these values into a similarity set. The \textit{overall similarity} is characterized by two statistical measures: mean and variance. The mean similarity represents the central tendency of the dataset’s feature similarity, with higher values indicating lower diversity due to greater redundancy among images. Meanwhile, the variance captures the dispersion of similarity values, reflecting the heterogeneity of the dataset. A lower variance suggests a more uniform distribution, whereas a higher variance indicates a greater mix of unique and similar samples. In our implementation, image features are extracted using the pre-trained VGG16 model \citep{simonyan2014very}. A similar approach is used to compute \textit{overall text similarity}, where text features are extracted using the pre-trained Sentence-BERT model \citep{reimers2019sentence}. This ensures a consistent methodology for evaluating both image and text redundancy across the dataset. The specifics of the \textit{overall image similarity}, along with the \textit{overall text similarity}, are provided in Appendix \ref{app:ImageSim} and \ref{app:TextSim}, respectively. 

\par Table \ref{tab:table3} presents a comparative analysis of our dataset with existing SAR datasets in terms of image similarity. In contrast to other SAR datasets \citep{shi2021object,li2024sardet}, our dataset demonstrates lower image similarity, indicating a higher degree of diversity and reduced redundancy among the SAR images in our collection. Furthermore, since existing SAR datasets lack text samples, we extend our comparison to include established RGB-text datasets, as detailed in Table \ref{tab:table3}. To ensure a fair evaluation of image similarity, we compute the similarity metric using a subset of RGB images derived from the RGB-SAR pairs in our dataset. This approach guarantees that the comparison is conducted within the same modality. The results reveal that the image similarity of our dataset is comparable to the mainstream remote sensing datasets containing RGB-text pairs.


\begin{table}[htbp]
\renewcommand{\arraystretch}{1.35}
    \centering
    \caption{\label{tab:table3}{\bf Comparison with Existing Remote Sensing Datasets in terms of Image and Text Similarity.} '\# Image' denotes the number of included images. 'Image Sim.' and 'Text Sim.' represent the similarity of image and text, respectively, which are represented as the form of 'mean/variance'. Lower mean or variance indicates higher dataset diversity. '--' means not available. }
    \scalebox{0.68}{
    \begin{tabular}{|c|c|c|c|c|}
        \hline
        \textbf{Modal} &\textbf{Dataset} & \textbf{\# Image} & \textbf{Image Sim.} & \textbf{Text Sim.} \\ 
        \hline
        
        \multirow{5}{*}{RGB} & RSICD \citep{lu2017exploring} & 10, 921 & 0.14 / 0.05& 0.27 / 0.16 
        \\ \cline{2-5}
        & RSITMD \citep{yuan2022exploring} & 4, 743 & 0.14 / 0.05& 0.20 / 0.13 \\ \cline{2-5}
        & UCM \citep{gao2021remote} & 2, 100 & 0.13 / 0.05& 0.22 / 0.17 \\ \cline{2-5}
        & VRSBench \citep{li2024vrsbench} & 29, 614 & 0.13 / 0.04& 0.48 / 0.12 \\ \cline{2-5}
        & Ours (paired RGB) & 13, 346 & 0.16 / 0.06 & -- \\ \hline
        \multirow{3}{*}{SAR} & FUSAR-Map \citep{shi2021object} & 610 & 0.28 / 0.09 & -- \\ \cline{2-5}
        & SARDet-100k \citep{li2024sardet} & 10, 4985 & 0.26 / 0.13 & -- \\ \cline{2-5}
        & Ours (SAR) & 118, 331 & 0.25 / 0.06& 0.49 / 0.16 \\ \hline
    \end{tabular}}
\end{table}  
\noindent\textbf{Supported Sub-tasks.} As illustrated in Figures \ref{fig:fig12}, the text annotations in our \textit{SARLANG-1M} dataset, including semantic captions and VQA labels, support seven applications: image description, object identification, object classification, instance counting, region referring, object positioning, and others. Table \ref{tab:table7} illustrates the relationship between dataset samples and each application, followed by the task definitions.


\textbf{1) Image Description:} This application involves structured parsing of satellite imagery by detailing key objects and their main characteristics. As shown in Table \ref{tab:table7}, the text annotations for image description primarily originate from the \textit{SARLANG-1M-Cap} benchmark.

\textbf{2) Object Identification:} This application evaluates the model’s SAR image recognition abilities by providing boolean answers to the presence of specific categories. The annotations mainly correspond to 'object identification' questions in the \textit{SARLANG-1M-VQA} benchmark.

\textbf{3) Object Classification:} This application focuses on identifying visible object categories within SAR images. The corresponding text annotations in our \textit{SARLANG-1M-VQA} benchmark encompass over 1,696 remote sensing object categories.


\textbf{4) Instance Counting:} For a specified object category, this application requires the model to quantify instances within the SAR image. 


\textbf{5) Region Referring:} Given a specific location in the SAR image, this application challenges the model to determine the category present in the local area. The 'region referring' samples in our \textit{SARLANG-1M-VQA} benchmark facilitate this application.

\textbf{6) Object Positioning:} This application assesses the model's capability to predict potential areas for specific object categories within SAR images. The 'object positioning' samples in our \textit{SARLANG-1M-VQA} benchmark offer precise regions for these categories.

\textbf{7) Others:} This category encompasses a variety of common tasks in the remote sensing domain, including object shape prediction, object direction prediction, land cover classification, pattern prediction, and reasoning functions. The 'general inquiries' samples in our \textit{SARLANG-1M-VQA} benchmark facilitate this application. Detailed examples for each function are shown in Figures \ref{fig:fig12}. 


\begin{table}[h]
\renewcommand{\arraystretch}{1.35}
    \centering
    \caption{\label{tab:table7}{\bf Seven SAR Applications supported by the SAR-text Samples in Our \textit{SARLANG-1M} Dataset.} ‘Cap’ and 'VQA' denote the \textit{SARLANG-1M-Cap} benchmark and \textit{SARLANG-1M-VQA} benchmark, respectively.}
    \scalebox{0.72}{
    \begin{tabular}{|l|c|c|}
        \hline
        \textbf{Application} & \textbf{\textit{SARLANG-1M} Dataset} & \textbf{\# SAR-text Sample} \\ \hline
        Image Description & Cap &  45,650 \\\hline 
        Object Identification & VQA ('object identification' type) &  484,620 \\ \hline
        Object Classification & VQA ('object classification' type) &  132,525 \\ \hline
        Instance Counting & VQA ('instance counting' type)  &  117,382 \\ \hline
        Region Referring & VQA ('region referring' type) &  221,450 \\\hline
        Object Positioning & VQA ('object positioning' type) &  106,171 \\\hline
        Others & VQA ('general inquiries' type) &  18,479 \\\hline
    \end{tabular}}
\end{table}

\par
\begin{figure*}[!t]
\centering
\includegraphics[width=6.5in]{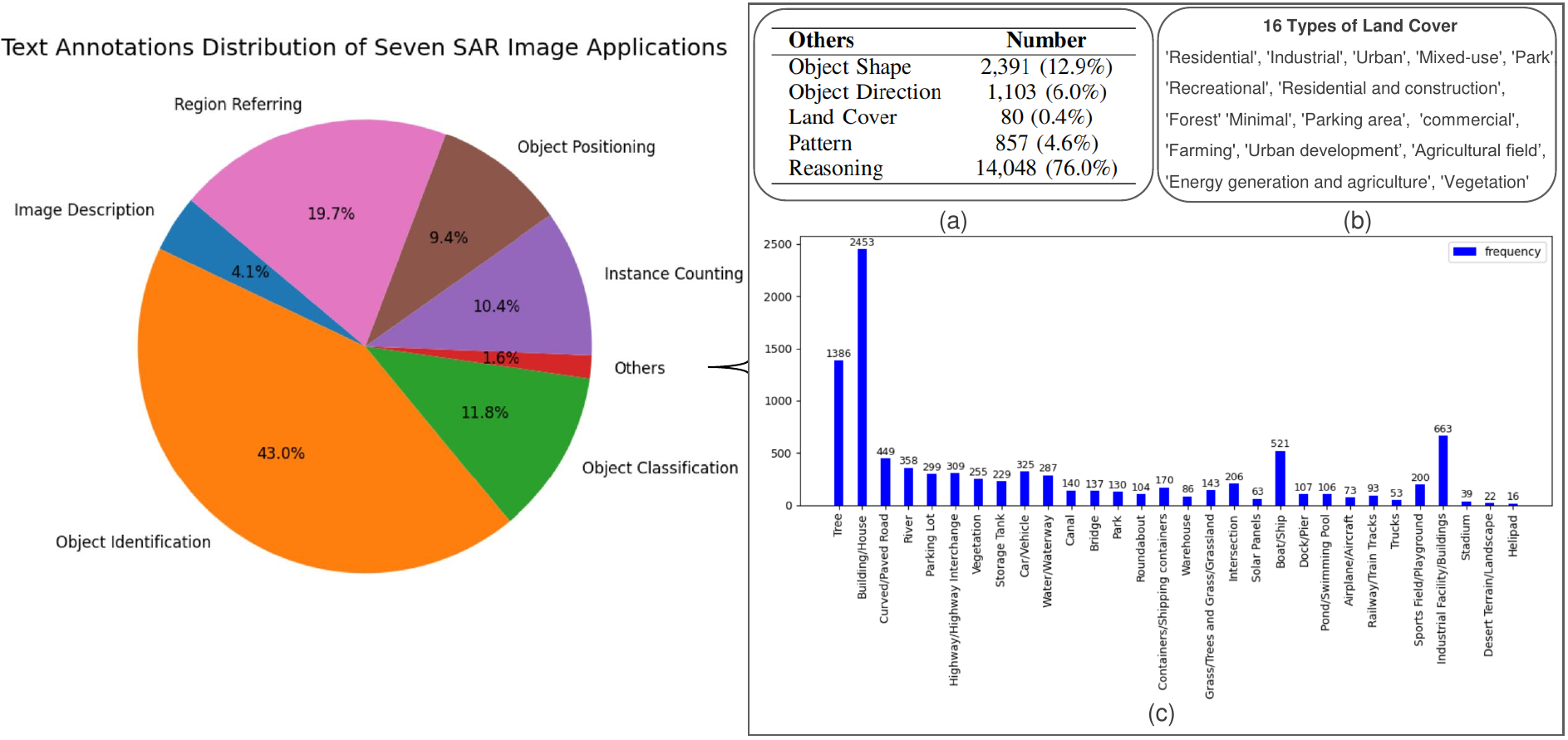}
\caption{\textbf{The Statistics of Text Annotations in \textit{SARLANG-1M} benchmark.} (a) Distribution of seven applications provided in \textit{SARLANG-1M} benchmark. (b) Numbers of each question types in the 'others' application. (c) Distribution of the 30 most frequent object categories.}
\label{fig:fig36}
\end{figure*}

 Our \textit{SARLANG-1M} dataset comprises a total of 1,126,2\-77 samples. The distribution of these samples across the seven applications is displayed on the left side of Figure \ref{fig:fig36}. Among these, the 'others' category contains a variety of text annotations, though they represent a smaller proportion. These annotations include five types of questions: object shape, direction, land cover, pattern, and reasoning. The statistics for each question type are listed in the right (a) side of Figure \ref{fig:fig36}. Notably, 'land cover' questions support a critical task in the remote sensing field, namely land cover classification. The 'land cover' item provides 80 text annotations and 16 land cover categories, as shown in the right (b) side of Figure \ref{fig:fig36}. Additionally, our \textit{SARLANG-1M} benchmark includes over 1,696 object categories. Figure \ref{fig:fig36} (c) displays the distribution of the 30 most frequent remote sensing categories. Some exhibit nuanced vocabulary in textual expressions. For instance, the 'Sports Field/Playground' category includes 'Soccer Field', 'Sports Field', 'Tennis Courts', 'Golf Course', and 'Baseball Field'. This diversity extends beyond the six categories in the object detection dataset \citep{li2024sardet}, enabling broader support for remote sensing applications.

\subsection{Dataset Construction}

\par In this section, we first provide a concise overview of the data sources that form the foundation of the \textit{SARLANG-1M} dataset. We then describe the dataset construction process, including SAR image preprocessing and the generation of high-quality textual annotations. Next, we outline the two different text generation pipelines employed for SAR image captioning and VQA tasks. Finally, we highlight key aspects of \textit{SARLANG-1M}, including its seven downstream applications and the statistical characteristics of its textual annotations.


\noindent\textbf{1) Data Source} 


\par The SAR images in \textit{SARLANG-1M} are sourced from four publicly available SAR datasets: SpaceNet6 \citep{shermeyer2020spacenet}, DFC2023 \citep{persello20232023}, OpenEarthMa\-p-SAR \citep{xia2025openearthmap}, and SARDet-100k \citep{li2024sardet}. Table \ref{tab:table1} provides a summary of these datasets, detailing their key characteristics and acquisition parameters.


\noindent \textbf{SpaceNet6 Dataset.} The SpaceNet6 dataset \citep{shermeyer2020spacenet} integrates half-meter quad-polarized X-band SAR images with corresponding half-meter optical imagery, focusing on the Port of Rotterdam, Netherlands. The SAR images was acquired using an aerial sensor operated by Capella Space, which captured the region over three consecutive days: August 4th, 23rd, and 24th, 2019. The dataset consists of 204 individual SAR image strips, each containing quad-polarized data (HH, HV, VH, and VV) in the X-band wavelength. The imagery was collected from an off-nadir perspective, with relative look angles ranging from 53.4° to 56.6°, and includes observations from both north- and south-facing acquisition directions.

\noindent \textbf{DFC2023 Dataset.} The DFC2023 dataset \citep{persello20232023} provides paired RGB and SAR images collected from multiple high-resolution satellites, including SuperView-1 (known as GaoJing in Chinese), Gaofen-2, and Gaofen-3, offering spatial resolutions of 0.5 m, 0.8 m, and 1 m, respectively. Additionally, Normalized Digital Surface Models (nDSMs) are provided as reference data, generated from stereo imagery captured by Gaofen-7 and WorldView satellites, with a ground sampling distance (GSD) of 2 m. The dataset spans seventeen cities across six continents, encompassing multi-resolution SAR images that enhances its geographic diversity.


\noindent \textbf{OpenEarthMap-SAR Dataset.} The OpenEarthMap-SAR dataset \citep{xia2025openearthmap} comprises 1.5 million segmented samples derived from 5,033 aerial and satellite images, each with a 1024×1024 pixel resolution. These images cover 35 regions across Japan, France, and the United States. The dataset includes a mix of partially manually annotated and fully pseudo-labeled eight-class land cover classifications, with a GSD ranging from 0.15 to 0.5 meters.
The SAR samples in OpenEarthMap-SAR are sourced from Umbra, acquired in Spotlight mode, with a resolution varying between 0.15 and 0.5 meters. The optical data contains red, green, and blue spectral bands, while the SAR data primarily consists of amplitude information in VV or HH polarization bands. To ensure precise alignment between optical and SAR datasets, multiple experts manually co-registered the paired images and conducted cross-verification to meet strict quality standards.


\noindent \textbf{SARDet-100k Dataset.} The SARDet-100k dataset \citep{li2024sardet} is a large-scale SAR object detection dataset, comprising 104,985 images spanning six major object categories: ship, tank, aircraft, bridge, car, and harbor. The dataset aggregates SAR images from ten publicly available datasets, captured by seven different satellites, including Gaofen-3, Sentinel-1, TerraSAR-X, and RadarSat-2 etc. Each SAR image in SARDet-100k is annotated with precise bounding boxes, formatted as $(category, [x\_min, y\_min, \newline width, height])$, where $x\_min$ and $y\_min$ denote the coordinates of the upper-left corner of the bounding box, while $category$ specifies the object category contained within the bounding box. $width$ and $height$ refer to the width and height of the bounding box in the SAR images. This dataset provides a rich resource for SAR-based object detection, with extensive annotations that enhance the training and evaluation of deep learning models.

\par These four publicly available datasets play distinct roles in constructing the \textit{SARLANG-1M} dataset. SpaceNet6 \citep{shermeyer2020spacenet}, DFC2023 \citep{persello20232023}, and OpenEarthMap-SAR \citep{xia2025openearthmap} contribute to both the \textit{SARLANG-1M-Cap} and \textit{SARLANG-1M-VQA} benchmarks, while SARDet-100k \citep{li2024sardet} is used exclusively for the \textit{SARLANG-1M-VQA} benchmark.
To ensure data integrity, SAR images with incomplete or missing information were removed from the original datasets. After this refinement, the \textit{SARLANG-1M-Cap} benchmark comprises 13,346 SAR images, while the \textit{SARLANG-1M-VQA} benchmark includes 118,331 SAR images.


\noindent\textbf{2) SAR Image Preprocessing} 

\par SAR imaging quality is often degraded by multiplicative speckle noise and artifacts. Following established SAR practices \citep{li2024sardet,qin2021multilevel}, we implement preprocessing to improve SAR image clarity and quality. 
\noindent \textbf{SARDet-100k Dataset.}
The SAR images in the SARDet-100k \citep{li2024sardet} dataset has been preprocessed and denoised. These SAR images have been cropped into 512*512 patches. Original SAR images are directly collected in our \textit{SARLANG-1M} dataset without any preprocessing operations.

\begin{figure}[!t]
\centering
\includegraphics[width=3.3in]{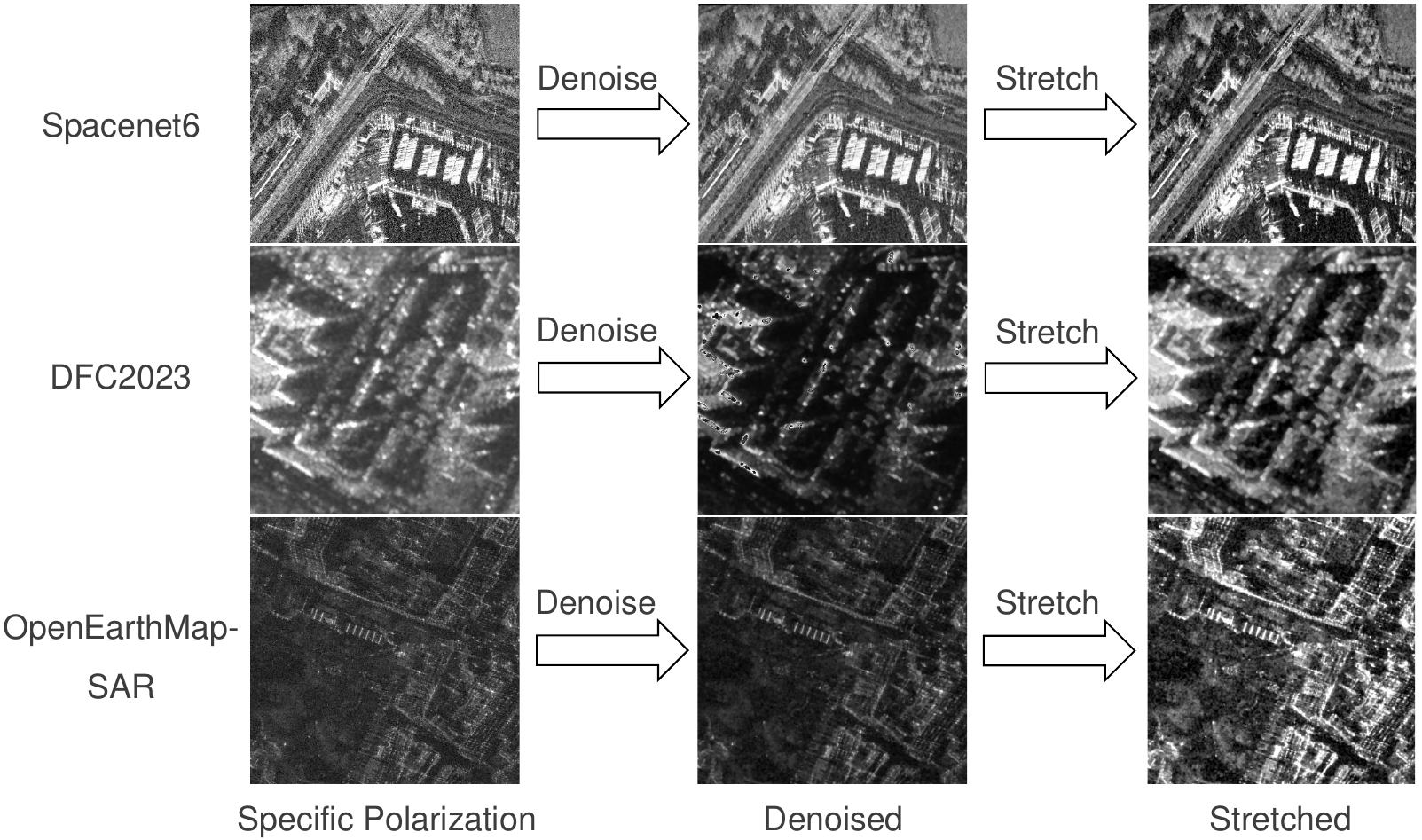}
\caption{\textbf{Pre-processing Pipeline for SAR Images in Our Benchmark.} All SAR images in our dataset, except those from the SARDet-100k \citep{li2024sardet} dataset, undergo a standardized preprocessing pipeline, which includes the selection of specific polarizations, followed by denoising \citep{yommy2015sar} and subsequent contrast enhancement \citep{ai2019outliers} to optimize image quality and interpretability.}
\label{fig:fig31}
\end{figure}

\noindent \textbf{SpaceNet6, DFC2023 and OpenEarthMap-SAR Dataset.} 
Addressing common issues such as low contrast and noise in SAR images from SpaceNet6 \citep{shermeyer2020spacenet}, DFC2023 \citep{persello20232023} and OpenEarthMap-SAR \citep{xia2025openearthmap} datasets, we apply single-channel transformation, denoising, and contrast stretching prior to VLM analysis, as illustrated in Figure \ref{fig:fig31}. For unified processing, specific polarized SAR images were selected as single-channel images. Note that, for single polarized SAR images in DFC2023 \citep{persello20232023} and OpenEarthMap-SAR \citep{xia2025openearthmap} datasets, the original band is selected as the final image channel. As for quad-polarized SAR images in SpaceNet6 \citep{shermeyer2020spacenet}, one of four bands (i.e. HH, VV, HV, and VH) is selected as the final image channel. These single-channel SAR images undergo further processing through denoising and contrast stretching. The refined Lee filter \citep{yommy2015sar} is utilized for denoising, while linear stretching \citep{ai2019outliers} is applied after logarithmic transformation for contrast enhancement. These preprocessing operations result in significantly enhanced image clarity and effectively highlight key objects within the SAR images.


\noindent\textbf{(3) Text Annotations Generation} 



\par We employ two distinct text generation strategies to create high-quality textual descriptions for the \textit{SARLANG-1M} benchmark, covering both \textit{SARLANG-1M-Cap} and \textit{SARLA\-NG-1M-VQA}. For \textit{SARLANG-1M-Cap}, we adopt a modality transfer approach, where textual descriptions are first generated for RGB images and then aligned with their corresponding SAR images. Since the paired RGB and SAR images depict the same content, this method enables the transfer of semantic information from the well-established RGB domain to SAR images.
For \textit{SARLANG-1M-VQA}, which primarily focuses on grounding and referring tasks for fine-grained SAR image understanding, we generate text descriptions directly from annotated bounding boxes in existing SAR datasets. This process constructs a novel text corpus, serving as a dedicated annotation resource for SAR VQA tasks.

\begin{figure*}[!t]
\centering
\includegraphics[width=6.8in]{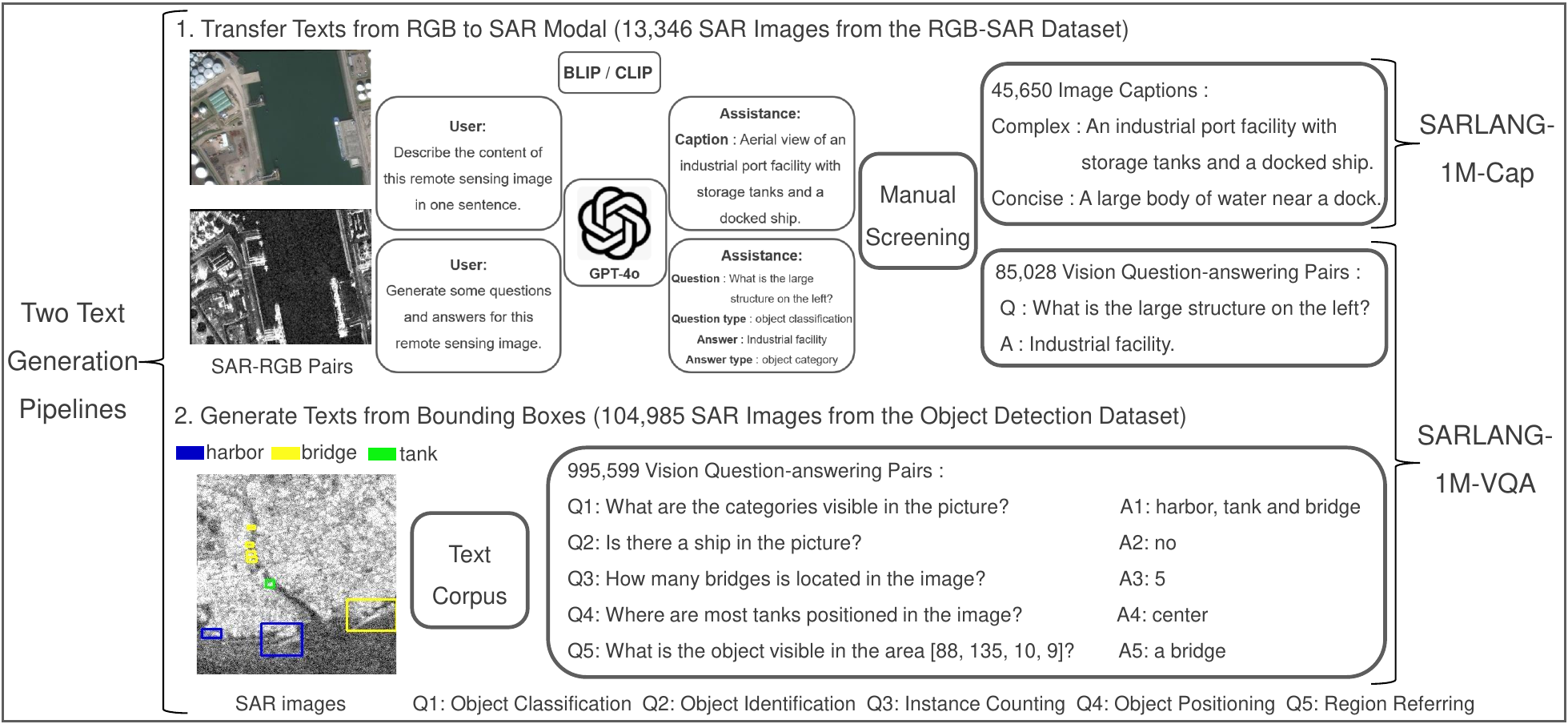}
\caption{\textbf{Text Annotations Generation Process of our \textit{SARLANG-1M} dataset, including \textit{SARLANG-1M-Cap} and \textit{SARLANG-1M-VQA} benchmark.} Two text generation pipelines are adopted to construct our \textit{SARLANG-1M} dataset. Notably, the manual screening is introduced to improve the quality of generated text through the supervisor of SAR experts. The text corpus is designed to generate text annotations from bounding boxes provided in the object detection dataset \citep{li2024sardet}. Objects in blue box, yellow box and green box refer to ‘harbor’, ‘bridge and ‘tank’ category, respectively. Q1, Q2, Q3, Q4, Q5 refer to the question type of object classification, object identification, instance counting, object positioning and region referring respectively.} 
\label{fig:fig32}
\end{figure*}
\begin{figure}[!t]
\centering
\includegraphics[width=3.2in]{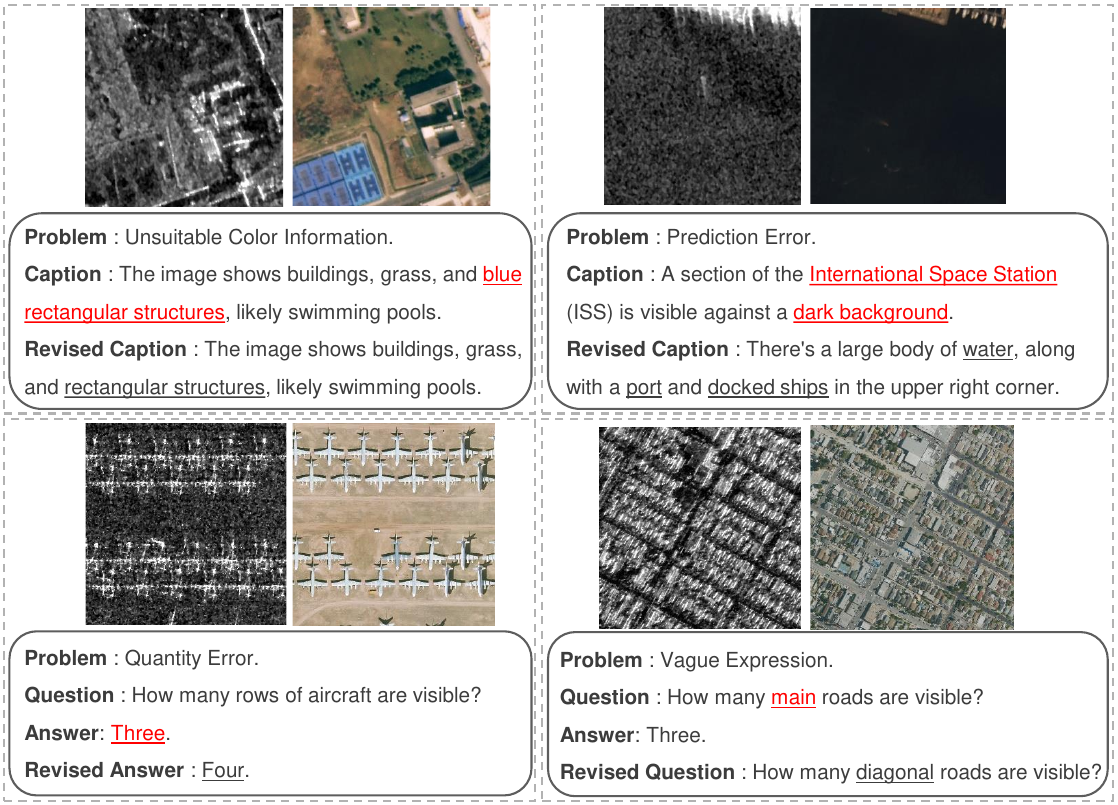}
\caption{\textbf{Exemplar False Cases before and after Manual Refining.} Four types of cases are presented: unsuitable color information, prediction errors, quantity errors, and vague expressions.} 
\label{fig:fig35}
\end{figure}

\begin{table*}[htbp]
\renewcommand{\arraystretch}{1.35}
\centering
\caption{\label{tab:table2}{\bf Text Corpus of \textit{SARLANG-1M-VQA} Dataset.} The templates of five question types and corresponding answers are listed.}
\scalebox{0.88}{
\begin{tabular}{|>{\centering\arraybackslash}p{3cm}|>{\centering\arraybackslash}p{11cm}|>{\centering\arraybackslash}p{4cm}|}
\hline
\textbf{Question Type} & \textbf{Question Template} & \textbf{Answer} \\ \hline
Object Identification & 
\begin{tabular}{@{}c@{}}
Does the/Are there [specific category] exist in the image? \\
Is there a [specific category] located on the [specific position] side of the image? \\
Are there any other categories apart from [specific category]/in the image?
\end{tabular} & 
‘Yes/No’ \\ \hline
Object Classification & 
What is the main object/are the categories visible in the image? & 
‘Ship’, ‘Aircraft’, ‘Car’, ‘Tank’, ‘Bridge’, ‘Harbor’ \\  \hline
Instance Counting & 
How many [specific category] are visible/is located in the image? & 
Number \\  \hline
Object Positioning & 
Where is/are the most [specific category] positioned within the image? & 
'Left', 'Right', 'Top', 'Bottom', 'Center' \\ \hline
Region Referring & 
What is the object in the area [detailed coordinates]? & 
‘Ship’, ‘Aircraft’, ‘Car’, ‘Tank’, ‘Bridge’, ‘Harbor’ \\ \hline
\end{tabular}}
\end{table*}

\noindent1) \textit{SARLANG-1M-Cap}

\par The SAR image captioning task involves generating a comprehensive textual description for a given SAR image, encapsulating intricate object details and contextual relationships. Unlike traditional classification tasks, this challenge requires predicting the global content of SAR images, ensuring that the generated captions capture not only the object categories present within the scene but also the spatial and semantic relationships among them. 

\noindent \textbf{Text generation.} \textit{SARLANG-1M-Cap} is designed to provide detailed descriptions of SAR images across multiple dimensions, including global scene context, object-level information, and inter-object relationships. To generate enriched and high-quality captions, we leverage three representative VLMs applied to paired RGB-SAR images:
\begin{itemize}[noitemsep,nolistsep]
    \item {BLIP \citep{li2022blip}: Utilizes ViT-Large/16 \citep{dosovitskiy2021vit} as its backbone and has been pretrained on 14 million images, including two human-annotated datasets (COCO \citep{lin2014microsoft} and Visual Genome \citep{krishna2017visual}) and three web-scale datasets (Conceptual Captions \citep{changpinyo2021conceptual}, Conceptual 12M \citep{changpinyo2021conceptual}, and SBU Captions \citep{ordonez2011im2text}).}
    \item {CLIP \citep{radford2021learning}: Initially pretrained on the large-scale LAION-2B dataset \citep{schuhmann2021laion}, followed by fine-tuning on MSCOCO \citep{lin2014microsoft}, making it well-suited for open-domain vision-language understanding.}
    \item {GPT-4o \citep{achiam2023gpt}: A state-of-the-art multimodal model capable of generating context-aware and detailed image captions, leveraging its advanced language modeling capabilities.}
\end{itemize}
As shown in Figure \ref{fig:fig32}, BLIP \citep{li2022blip} and CLIP \citep{radford2021learning} primarily generate concise captions, whereas GPT-4o \citep{achiam2023gpt} produces more complex and detailed descriptions. This diverse captioning approach enhances sentence richness and ensures that the generated text is not confined to a rigid or repetitive linguistic pattern.

\noindent2) \textit{SARLANG-1M-VQA}

\par Given a SAR image, the VQA task involves answering questions based on the semantic understanding of the image content, as illustrated in Figure \ref{fig:fig32}. While \textit{SARLANG-1M-Cap} emphasizes global-level understanding, \textit{SARLANG-1M-VQA} is designed to place greater focus on local content comprehension—including the identification, localization, and reasoning about specific objects and regions within the SAR images.

\textbf{Text generation.} To produce high-quality text annotations that provides precise quantifications and coordinate specifications for visible objects in SAR images, we introduce a novel text corpus as shown in Table \ref{tab:table2}. Five primary question templates and corresponding answers are defined:
\begin{itemize}[noitemsep,nolistsep]
    \item {Object Identification: Designed to determine the presence of specific objects such as 'Ship', 'Tank', 'Aircraft', 'Bridge', 'Car', and 'Harbor', with answers being 'yes' or 'no'.}
    \item {Object Classification: Aims to identify the predominant category within the SAR image, with answers from the set ['Ship', 'Tank', 'Aircraft', 'Bridge', 'Car', 'Harbor'].}
    \item {Instance Counting: Investigates the quantity of a specific category within the SAR image, with the count as the answer.}
    \item {Object Positioning: Determines the approximate location of a category, with answers being one of ['Left', 'Right', 'Top', 'Bottom', 'Center'].}
    \item {Region Referring: Focuses on identifying the category within a specified area, with answers from the six potential categories ['Ship', 'Tank', 'Aircraft', 'Bridge', 'Car', 'Harbor'].}
\end{itemize}

\par To further diversify question-answer patterns and expand remote sensing categories beyond those in the SAR object detection dataset \citep{li2024sardet}, we apply a similar modality transfer approach to generate VQA annotations. As shown in Table \ref{fig:fig32}, various prompts are input to the GPT-4o \citep{achiam2023gpt} model, producing a wide range of question-answer pairs and enhancing interpretative depth. In addition to enriching the text annotations for the five defined question types in our corpus, the remaining VQA labels constitute an 'general inquiries' question type. This type of questions explores novel remote sensing applications such as land cover classification, reasoning, and object shape prediction, as shown in Table. \ref{fig:fig12}.

\noindent {3) Quality Control} 

\par While the automated text generation pipeline provides a scalable approach to annotating SAR images, it is not without limitations. First, SAR images inherently lack color information, leading to misalignment when descriptions include color-based attributes derived from paired RGB images. Second, the generated text annotations are not always accurate, as current VLMs have limited capability in fully comprehending remote sensing RGB imagery, which can result in errors when transferring descriptions to SAR images.
To ensure high-quality textual annotations in \textit{SARLANG-1M}, we implement a rigorous manual review and filtering process, conducted by domain experts. This ensures that incorrect, inconsistent, or irrelevant descriptions are identified and revised. Figure \ref{fig:fig35} presents examples of failure cases and their corresponding corrections after expert verification.
 
\section{Experimental Analysis}
\label{sec:experiments}
\par In this section, some mainstream VLMs and non-VLMs are adopted to comprehensively evaluate our constructed \textit{SARLANG-1M} dataset and demonstrate the effectiveness of our dataset. First, the experiment setup, including implementation details and evaluation metrics, is introduced. Then, extensive experiments on SAR image captioning and VQA tasks are conducted on our \textit{SARLANG-1M} dataset with detailed analyzing. We also provide some visual results to present the text quality of our dataset more intuitively. Finally, we provide ablation analyses on the SAR image preprocessing strategies proposed in this paper and justify their contributions in SAR image captioning and VQA tasks.

\subsection{Experiment Setup}

\subsubsection{Implementation Details}

\par In order to facilitate benchmark evaluation, the \textit{SARLAN\-G-1M} dataset has been divided into two distinct, non-overlapping subsets, each specifically curated for model training and evaluation. The SAR-text pairs distribution statistics in our dataset are shown in Table \ref{tab:table9}. For the \textit{SARLANG-1M-Cap} benchmark, following standard methodologies common in remote sensing image datasets \citep{xia2023openearthmap, li2024vrsbench}, all SAR-text samples are divided into training and test sets using a 7:3 ratio. This division resulted in a training set comprising 9,341 images paired with 31,968 captions and a test set consisting of 4,005 images accompanied by 13,682 captions, ensuring a robust foundation for VLMs training and evaluation. 

\begin{table}[h]
\renewcommand{\arraystretch}{1.35}
    \centering
\caption{\label{tab:table9}{\bf SAR-text Pairs Distribution Statistics in Our \textit{SARLANG-1M} dataset.}}
    \scalebox{0.9}{
    \begin{tabular}{|l|c|c|c|}
        \hline
        \textbf{Benchmark} &
        \textbf{Text} & \textbf{\# Train} & \textbf{\# Test} \\ \hline
        \multirow{2}{*}{\textit{SARLANG-1M-Cap}} & Complex Caption & 14,481  & 6,190  \\ 
        & Concise Caption & 17,487 & 7,492 \\ \cline{1-4}
        \textit{SARLANG-1M-VQA} & VQA & 955,372  & 125,255 \\ \hline
    \end{tabular}}
\end{table}

\par In the \textit{SARLANG-1M-VQA} benchmark, for SAR samples identical to those in the \textit{SARLANG-1M-Cap} benchmark, the same 7:3 ratio partitioning method for training and test sets has been employed. For the SAR images originating from the dataset \citep{li2024sardet}, we preserved the original train/test image splits as established in the SARDet-100k dataset \citep{li2024sardet}. Consequently, the training component encompasses 103,834 SAR images paired with 955,372 corresponding textual annotations, while the evaluation subset includes 14,497 SAR images with 125,255 associated text annotations. 


\par For the aforementioned benchmarks, we evaluate ten mainstream VLMs to demonstrate their potential in understanding SAR images and to highlight the contribution of our \textit{SARLANG-1M} dataset. To ensure a fair and comprehensive evaluation, we employ two distinct evaluation methodologies for these VLMs.
First, we initialize the models with pre-trained parameters from large-scale image-text alignment datasets and directly evaluate their performance on the test set of the \textit{SARLANG-1M} benchmark. This approach allows for an intuitive comparison of the VLMs' capabilities in SAR downstream tasks without additional fine-tuning.
Second, we fine-tune each VLM using the training set of the \textit{SARLANG-1M} dataset and subsequently evaluate their performance on the test set of the \textit{SARLANG-1M} benchmark. By comparing the results before and after fine-tuning, we quantitatively assess the contribution of the \textit{SARLANG-1M} dataset to improving model performance on SAR downstream tasks.
Among the evaluated VLMs, five models—BLIP \citep{li2022blip}, LLaVA1.5-7B \citep{liu2024visual}, LLaVA1.5-13B \citep{liu2024visual}, QWEN2-VL-7B \citep{wang2024qwen2} and QWEN2.5-VL-7B \citep{Qwen2.5-VL}—are fine-tuned using the Low-Rank Adaptation (LoRA) \citep{hu2022lora} training method with a rank of 8 applied to all linear layers. This fine-tuning strategy ensures efficient adaptation while minimizing computational overhead. During the fine-tuning stage, each VLM is trained for 3 epochs with a batch size of 1. The learning rate is initialized at 1e-4 and incorporates a warm-up strategy with a ratio of 0.1. During the evaluation of these VLMs, the prompt for the SAR image captioning task is: "Describe the content of this image in one sentence," while the prompt for the SAR image VQA task is: "Question: \{question\}. Answer the question concisely."


\par For the SAR image captioning task, two traditional models, namely MLAT \citep{liu2022remote} and HCNet \citep{yang2024hcnet}, are utilized as baseline networks to provide the comparison with VLMs. These traditional models are trained from scratch, adhering to their original specifications and hyperparameters. Additionally, they follow the same train/test split as the VLMs. All experiments are performed on a single NVIDIA A100 GPU. Our source code is publicly available to facilitate community replication and to support ongoing research in {\url{https://github.com/Jimmyxichen/SARLANG-1M}}.


\subsubsection{Evaluation Metrics}

\par For SAR image captioning model evaluation, we adhere to standard practices established by existing VLMs \citep{li2024vrsbench}, utilizing a comprehensive set of established metrics including BLEU \citep{papineni2002bleu}, ROUGE\_L \citep{lin2004rouge}, and CIDEr \citep{vedantam2015cider}. For the BLEU metric, we consider n-gram precision with n values of 1, 2, 3, and 4. A higher BLEU score indicates better performance and higher-quality text outputs. 



\par For the evaluation of SAR image VQA tasks, we employ a GPT-4 based metric overall accuracy \citep{li2024vrsbench} to assess the performance of existing VLMs on the \textit{SARLANG-1M-VQA} benchmarks. In the GPT-4 based evaluation, we utilize GPT-4 to determine whether the predicted answers match the ground truth for each question. The prompt used is: “Question: \{question\}, Ground Truth Answer: \{ground\_truth\}, Predicted Answer: \{predicted answer\}. Does the predicted answer match the ground truth? Answer 1 for match and 0 for not match. Use semantic meaning, not exact match. Synonyms are also treated as a match, e.g., oval and circular, pond and swimming pool.” Overall accuracy is calculated as the ratio of the number of matches (1s) to the total number of questions.

\begin{table*}[ht]
\renewcommand{\arraystretch}{1.35}
\centering
\caption{\label{tab:table4}{\bf Comparison with Existing Image Captioning Methods on \textit{SARLANG-1M-Cap} Benchmark.} The complex text and concise text refer to ground truth generated by GPT-4o \citep{achiam2023gpt} model and BLIP \citep{li2022blip}/CLIP \citep{radford2021learning} models, respectively.}
\scalebox{0.81}{
\begin{tabular}{|l|l|l|l|c|c|c|c|c|c|}
\hline
\textbf{Caption} & \textbf{Model}& \textbf{Param} & \textbf{Train Strategy} & \textbf{BLEU\_1} & \textbf{BLEU\_2} & \textbf{BLEU\_3} & \textbf{BLEU\_4} & \textbf{ROUGE\_L} & \textbf{CIDEr} \\ \hline
\multirow{16}{*}{Complex} 
& MLAT \citep{liu2022remote} & \textbf{--} & Train from Scratch & 26.02 & 16.90 & 11.67 & 8.09  & 24.90 & 34.50 \\ 
& HCNet \citep{yang2024hcnet} & \textbf{--} & Train from Scratch & 26.31 & 16.57 & 11.12 & 7.42 & 24.91 & 33.61 \\ \cline{2-10} 
& BLIP(Vit-Base) \citep{li2022blip} & 129M  & LoRA Finetune & 26.32 & 17.31 & 12.09& 8.56 & 24.63 & 36.54 \\ 
& LLaVA1.5 \citep{liu2024visual} & 13B & LoRA Finetune & 34.90& 22.95& 16.55& 12.01 & 32.43 & 45.13  \\ 
& LLaVA1.5 \citep{liu2024visual} & 7B & LoRA Finetune & 35.24 & 23.63&  17.28& 12.70 & 32.72 & 46.35  \\ 
& QWEN2-VL \citep{wang2024qwen2} & 7B & LoRA Finetune & \textbf{35.78} & \textbf{23.72} & \textbf{17.57} & \textbf{13.08}  & \textbf{32.84} & 48.36 \\ 
& QWEN2.5-VL \citep{Qwen2.5-VL} & 7B & LoRA Finetune & 32.79 & 22.29 & 16.44 & 12.25  & 30.24 &\textbf{55.64} \\ 
\cline{2-10}
& LLaVA1.5 \citep{liu2024visual} & 13B & Without Finetune & 6.69 & 2.88 & 1.11 & 0.44  & 11.50 & 0.01 \\ 
& LLaVA1.5 \citep{liu2024visual} & 7B & Without Finetune & 7.13 & 3.08 & 1.11 & 0.44  & 12.06 & 0.03 \\ 
& DeepSeek-VL \citep{lu2024deepseek} & 7B & Without Finetune & 21.42 & 9.58& 4.17 & 2.19  & 19.17 & 4.62 \\ 
& DeepSeek-VL \citep{lu2024deepseek} & 1.3B & Without Finetune & 22.35 & 8.23& 3.69 & 1.89  & 18.88 & 5.80 \\ 
& InternVL2.5 \citep{chen2024internvl} & 8B & Without Finetune & 28.11 & 17.90 & 12.19 & 7.51  & 29.66 & 16.15 \\ 
& InternVL2.5 \citep{chen2024internvl} & 4B & Without Finetune & 27.96 & 18.28 & 12.60 & 7.89  & 29.98 & 15.38 \\ 
& QWEN2-VL \citep{wang2024qwen2} & 7B & Without Finetune & 6.48 & 2.78 & 0.96 & 0.35  & 11.33 & 0.01 \\ 
& QWEN2.5-VL \citep{Qwen2.5-VL} & 7B & Without Finetune & 24.56 & 13.69 & 8.87 & 5.35  & 24.49 & 9.42 \\ 
& QWEN2.5-VL \citep{Qwen2.5-VL} & 3B & Without Finetune & 26.70 & 15.31 & 9.04 & 4.14  & 26.06 & 8.96 \\ 
\hline
\multirow{16}{*}{Concise} 
& MLAT \citep{liu2022remote} & \textbf{--} & Train from Scratch &  60.55& 53.82& 48.46& 43.33  & 60.03 & 224.07  \\ 
& HCNet \citep{yang2024hcnet}& \textbf{--} & Train from Scratch & \textbf{61.06} & \textbf{54.45} & \textbf{49.19} & \textbf{44.15} & \textbf{60.78} & \textbf{239.00}  \\ \cline{2-10} 
& BLIP(Vit-Base) \citep{li2022blip} & 129M  & LoRA Finetune & 58.34 & 51.98& 46.95 & 42.16  & 55.73 & 227.24 \\ 
& LLaVA1.5 \citep{liu2024visual} & 13B & LoRA Finetune &  25.42 & 15.61 & 10.64 & 7.20  & 22.65 & 11.03 \\ 
& LLaVA1.5 \citep{liu2024visual} & 7B & LoRA Finetune &  23.97 & 14.09 & 9.00 & 5.71  & 21.52 & 10.27\\ 
& QWEN2-VL \citep{wang2024qwen2} & 7B & LoRA Finetune &  24.58 & 15.14 & 10.41 & 7.11  & 22.28 & 9.56\\
& QWEN2.5-VL \citep{Qwen2.5-VL} & 7B & LoRA Finetune & 37.75 & 31.00 & 26.31 & 22.36  & 38.50 & 70.05 \\ 
\cline{2-10}
& LLaVA1.5 \citep{liu2024visual} & 13B & Without Finetune & 9.23 & 5.12 & 2.93 & 1.73  & 15.08 & 0.01 \\ 
& LLaVA1.5 \citep{liu2024visual} & 7B & Without Finetune &  9.22 & 4.84 & 2.26 & 1.07  & 14.72 & 0.02  \\
& DeepSeek-VL \citep{lu2024deepseek} & 7B & Without Finetune &  20.15 & 10.50 & 5.12 & 3.01  & 17.61 & 2.73  \\
& DeepSeek-VL \citep{lu2024deepseek} & 1.3B & Without Finetune &  24.98 & 14.88 & 6.33 & 3.45  & 22.60 & 3.78  \\
& InternVL2.5 \citep{chen2024internvl} & 8B & Without Finetune &  29.50 & 20.72 & 14.93 & 9.85  & 30.28 & 4.84  \\
& InternVL2.5 \citep{chen2024internvl} & 4B & Without Finetune &  26.58 & 17.50 & 12.02 & 7.47  & 26.89 & 4.46  \\
& QWEN2-VL \citep{wang2024qwen2} & 7B & Without Finetune &  8.38 & 4.27 & 2.07 & 1.02  & 14.07 & 0.01  \\
& QWEN2.5-VL \citep{Qwen2.5-VL} & 7B & Without Finetune & 18.42 & 7.85 & 3.90 & 1.57  & 17.68 & 2.85 \\ 
& QWEN2.5-VL \citep{Qwen2.5-VL} & 3B & Without Finetune & 32.17 & 23.10 & 17.33 & 12.17  & 30.97 & 5.48 \\ 
\hline
\end{tabular}}
\end{table*}
\begin{table}[ht]
\renewcommand{\arraystretch}{1.35}
\centering
\caption{\label{tab:table8}{\bf Comparison with Existing Image VQA Methods on \textit{SARLANG-1M-VQA} Benchmark.} The accuracy metric denotes the ratio of correctly answered questions. }
\scalebox{0.75}{
\begin{tabular}{|l|l|l|c|}
\hline
\textbf{Model} & \textbf{Param} & \textbf{Train Strategy}  & \textbf{Accuracy} \\ \hline
LLaVA1.5 \citep{liu2024visual} & 13B & Without Finetune  & 29.91 \\ 
LLaVA1.5 \citep{liu2024visual} & 7B & Without Finetune & 33.55  \\ 
QWEN2-VL \citep{wang2024qwen2} & 7B & Without Finetune  & 28.33 \\
DeepSeek-VL \citep{lu2024deepseek} & 7B & Without Finetune &  29.26  \\
DeepSeek-VL \citep{lu2024deepseek} & 1.3B & Without Finetune &  30.02  \\
InternVL2.5 \citep{chen2024internvl} & 8B & Without Finetune &  39.90  \\
InternVL2.5 \citep{chen2024internvl} & 4B & Without Finetune &  \textbf{43.95}  \\
QWEN2.5-VL \citep{Qwen2.5-VL} & 7B & Without Finetune & 38.60  \\ 
QWEN2.5-VL \citep{Qwen2.5-VL} & 3B & Without Finetune & 41.23  \\ 
\hline 
LLaVA1.5 \citep{liu2024visual} & 13B  & LoRA Finetune  & 70.04 \\ 
LLaVA1.5 \citep{liu2024visual} & 7B & LoRA Finetune  & 70.30 \\ 
QWEN2-VL \citep{wang2024qwen2} & 7B & LoRA Finetune  & 68.55 \\
QWEN2.5-VL \citep{Qwen2.5-VL} & 7B & LoRA Finetune  & \textbf{73.33} \\
\hline
\end{tabular}}
\end{table}
\begin{table}[ht]
\renewcommand{\arraystretch}{1.35}
\centering
\caption{\label{tab:table82}{\bf Comparison between Human Evaluation and Performance of VLMs on {Validation Set} (30 samples) of \textit{SARLANG-1M-VQA} Benchmark.} The accuracy metric denotes the ratio of correctly answered questions. '--' refers to not applicable.}
\scalebox{0.75}{
\begin{tabular}{|l|l|l|c|}
\hline
\textbf{Model} & \textbf{Param} & \textbf{Train Strategy}  & \textbf{Accuracy} \\ \hline
QWEN2-VL \citep{wang2024qwen2} & 7B & Without Finetune  & 33.33 \\
DeepSeek-VL \citep{lu2024deepseek} & 7B & Without Finetune &  26.66  \\
DeepSeek-VL \citep{lu2024deepseek} & 1.3B & Without Finetune &  33.33  \\
InternVL2.5 \citep{chen2024internvl} & 8B & Without Finetune &  50.00  \\
InternVL2.5 \citep{chen2024internvl} & 4B & Without Finetune &  \textbf{53.33}  \\
QWEN2.5-VL \citep{Qwen2.5-VL} & 7B & Without Finetune & 36.66  \\ 
QWEN2.5-VL \citep{Qwen2.5-VL} & 3B & Without Finetune & 40.00  \\ 
\hline 
QWEN2-VL \citep{wang2024qwen2} & 7B & LoRA Finetune  & 56.66 \\
QWEN2.5-VL \citep{Qwen2.5-VL} & 7B & LoRA Finetune  & \textbf{63.33} \\
\hline 
Ordinary People & -- & --  & 27.76 \\
SAR Experts & -- & --  & \textbf{57.76} \\\hline 
\end{tabular}}
\end{table}

\begin{figure*}[!t]
\centering
\includegraphics[width=6.8in]{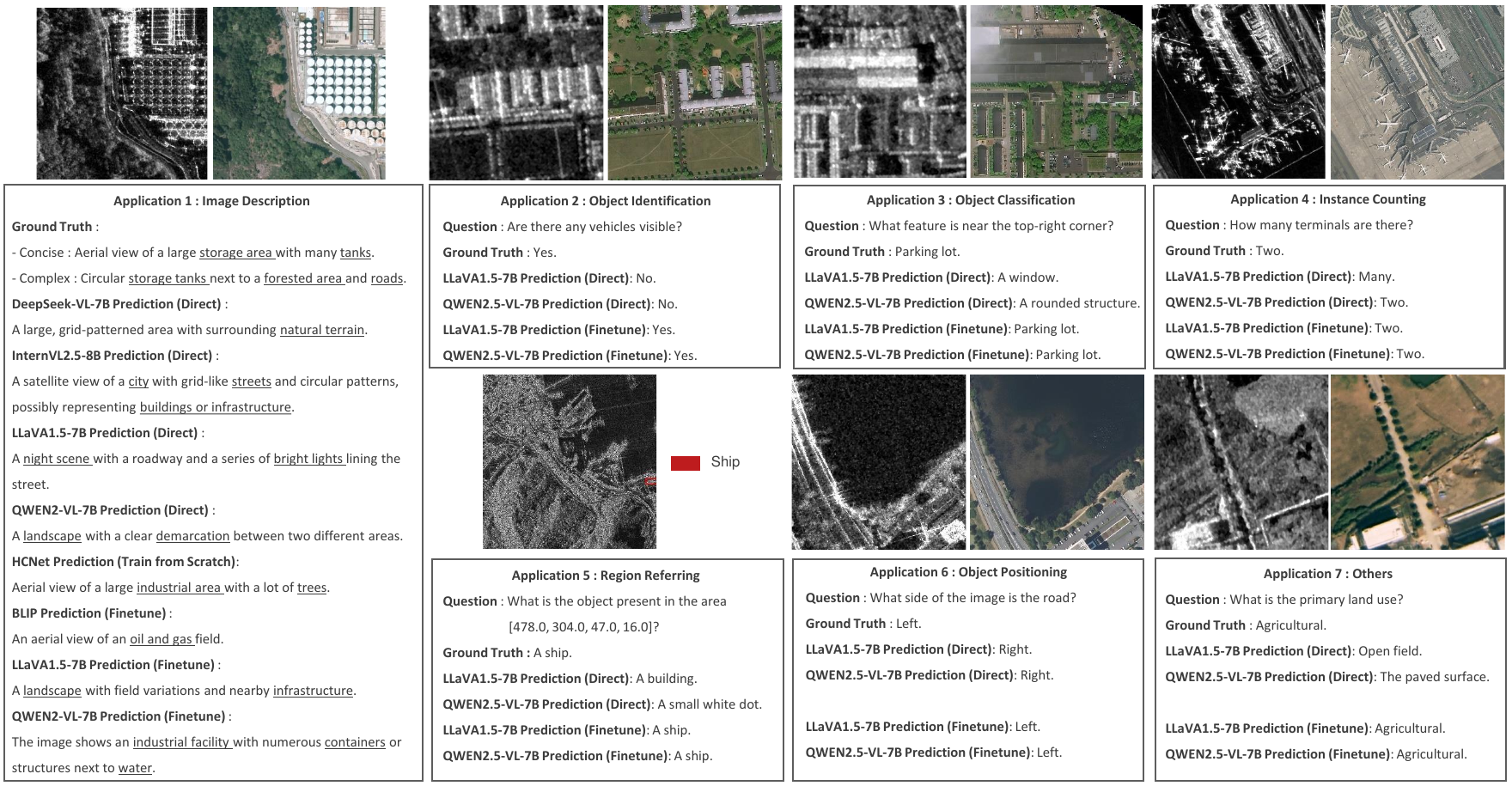}
\caption{\textbf{Visualization Results of Multiple VLMs' Predictions on our \textit{SARLANG-1M} benchmark} The visualization results are presented across seven distinct application scenarios. 'Direct' denotes evaluation without fine-tuning, while 'Fine-tune' represents evaluation after model fine-tuning on the \textit{SARLANG-1M} dataset. 'Train from Scratch' means evaluation after model training from scratch.}
\label{fig:fig38}
\end{figure*}
\begin{figure*}[!t]
\centering
\includegraphics[width=6in]{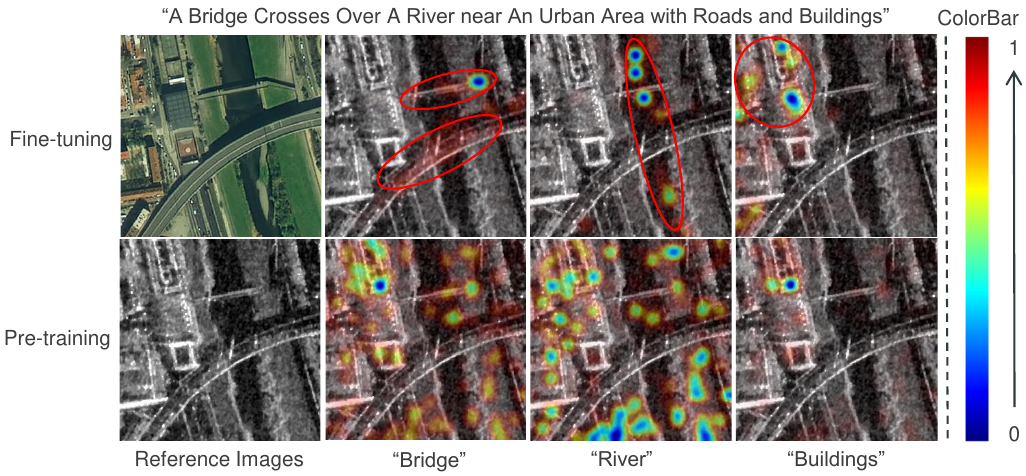}
\caption{\textbf{Per-word Grad-CAM Visualizations predicted by BLIP Model before and after Fine-tuning on our \textit{SARLANG-1M-Cap} Benchmark.} The text predicted by the BLIP \citep{li2022blip} model, along with the key targets identified within the text, is marked above and below the image, respectively. 'Pre-training' and 'Fine-tuning' denote the prediction results generated by the BLIP model before and after fine-tuning on our \textit{SARLANG-1M-Cap} dataset, respectively. Regions with strong responses to the given words are highlighted in red, while those with weak responses are highlighted in blue. After fine-tuning on our \textit{SARLANG-1M-Cap} dataset, the BLIP \citep{li2022blip} model improves in identifying areas corresponding to the given words, particularly those enclosed in red circles.}
\label{fig:fig39}
\end{figure*}

\subsection{Benchmarking \textit{SARLANG-1M}}

\subsubsection{SAR Image Captioning on \textit{SARLANG-1M-Cap} Benchmark}
\par As illustrated in Table \ref{tab:table4}, the InternVL2.5-8B \citep{chen2024internvl} model demonstrates significantly superior performance compared to other VLMs when evaluated on the test set without fine-tuning, achieving state-of-the-art results when evaluated against both complex and concise caption ground truths. This exceptional performance can be attributed to InternVL2.5-8B \citep{chen2024internvl}'s extensive pretraining on larger and more diverse datasets, which endows it with a comprehensive understanding of SAR image scenes and robust language generation capabilities. These advantages enable the InternVL2.5-8B \citep{chen2024internvl} model to produce more descriptive and accurate captions.

\par After fine-tuning on the \textit{SARLANG-1M-Cap} benchmark, the QWEN2-VL-7B \citep{wang2024qwen2} model achieves the highest performance metrics when evaluated against complex caption ground truth, with a BLEU\_1 score of 35.78, a ROUGE\_L score of 32.84, and a CIDEr score of 48.36. However, QWEN2-VL-7B \citep{wang2024qwen2} underperforms relative to traditional models when evaluated against the concise caption ground truth, suggesting that the outputs of traditional models align more closely with simple, fixed-pattern descriptions. Specifically, the HCNet \citep{yang2024hcnet} model achieves the highest performance among traditional models, with a BLEU\_1 score of 61.06, a ROUGE\_L score of 60.78, and a CIDEr score of 239.00. The MLAT \citep{liu2022remote} model follows closely, securing the second-highest performance with a BLEU\_1 score of 60.55, a ROUGE\_L score of 60.03, and a CIDEr score of 224.00. Notably, the BLIP \citep{li2022blip} model achieves the highest performance among all VLMs when evaluated against concise caption ground truth. This is primarily because most of the concise ground truth captions in the \textit{SARLANG-1M-Cap} benchmark are generated by the BLIP \citep{li2022blip} model, resulting in a closer alignment between its outputs and the benchmark's concise caption patterns.

\par Furthermore, the performance improvements observed after fine-tuning on the \textit{SARLANG-1M-Cap} benchmark validate the effectiveness of our dataset. Specifically, four VLMs—LLaVA1.5-7B \citep{liu2024visual}, LLaVA1.5-13B \citep{liu2024visual}, QWEN2-VL-7B \citep{wang2024qwen2} and QWEN2.5-VL-7B \citep{Qwen2.5-VL}—exhibit significant enhancements in performance metrics. Among these, QWEN2-VL-7B \citep{wang2024qwen2} demonstrates the most substantial improvements for the complex captions, with enhanced scores of 29.30 for BLEU\_1, 21.51 for ROUGE\_L, and 48.35 for CIDEr. As for the concise captions, QWEN2\-.5-VL-7B \citep{Qwen2.5-VL} achieves the highest performance improvements, with enhanced scores of 19.33 for BLEU\_1, 20.82 for ROUGE\_L, and 67.20 for CIDEr. These results underscore the value of the \textit{SARLANG-1M-Cap} benchmark in advancing the VLM's capabilities in SAR image captioning task.

\par To offer an intuitive comparison, we present a visualization of selected image captioning results from the \textit{SARLANG-1M-Cap} benchmark in Figure \ref{fig:fig38}. Notably, when evaluating the pre-trained LLaVA1.5-7B \citep{liu2024visual} model on the \textit{SARLANG-1M-Cap} benchmark, the model misclassifies the scene as "a night scene" and incorrectly identifies the key objects "storage tanks" as "bright lights." This misclassification arises because the pre-trained LLaVA\-1.5-7B \citep{liu2024visual} model lacks sufficient knowledge of SAR image. However, after fine-tuning on the \textit{SARLANG-1M-Cap} benchmark, the LLaVA1.5-7B \citep{liu2024visual} model demonstrates significant improvement, accurately identifying the scene and generating captions that include the correct keyword "infrastructure." Additionally, the QWEN2-VL-7B \citep{wang2024qwen2} model demonstrates comparable improvements, accurately identifying key elements such as "industrial facility" and "containers" within the SAR image. These results highlight the effectiveness of the \textit{SARLANG-1M-Cap} benchmark in enhancing the VLM's capability to interpret SAR image. Besides, we compare the text captions generated by HCNet \citep{yang2024hcnet}, BLIP \citep{li2022blip}, DeepSeek-VL-7B \citep{lu2024deepseek}, and InternVL2.5-8B \citep{chen2024internvl}. The HCNet \citep{yang2024hcnet} model demonstrates the ability to generate captions that accurately incorporate key terms such as "industrial area" and "trees." Similarly, the BLIP \citep{li2022blip} model produces captions containing the correct phrase "oil and gas field". In contrast, the DeepSeek-VL-7B \citep{lu2024deepseek} and InternVL2.5-8B \citep{chen2024internvl} models exhibit misclassifications, describing storage tanks as a "grid-patterned area" and "grid-like streets", respectively. These results reveal that models fine-tuned on the \textit{SARLANG-1M} dataset consistently generate more accurate and detailed captions. In comparison, models without fine-tuning on \textit{SARLANG-1M} tend to produce captions that include incorrect or ambiguous terminology. 

\par Furthermore, to investigate the correlation between the generated texts and SAR images, we constructed a per-word Grad-CAM \citep{selvaraju2017grad} visualization to illustrate the response levels between key vocabulary in the captions generated by the BLIP \citep{li2022blip} model  and key objects in the SAR images. In comparison with the results depicted in the second row in the Figure \ref{fig:fig39}, the results presented in the first row demonstrate that the \textit{SARLANG-1M-Cap} dataset significantly enhances the BLIP \citep{li2022blip} model's capability to accurately identify and localize key objects within SAR images.

\subsubsection{SAR VQA on \textit{SARLANG-1M-VQA} benchmark}


\par Table \ref{tab:table8} presents the VQA performance of various VLMs on our \textit{SARLANG-1M-VQA} dataset. As demonstrated in Table \ref{tab:table8}, among the models evaluated without fine-tuning on our \textit{SARLANG-1M-VQA} training set, the InternVL2.5-4B \citep{chen2024internvl} model achieves state-of-the-art performance, attaining an overall accuracy of 43.95\%. Besides, the QWEN2.5-VL-7B \citep{Qwen2.5-VL} model, after fine-tuning on the \textit{SARLANG-1M-VQA} training set, achieves the highest overall accuracy of 73.33\%. Furthermore, the baseline LLaVA1.5-7B \citep{liu2024visual} model with finetuning on our \textit{SARLANG-1M-VQA} training set gets an overall accuracy of 70.30\% on our \textit{SARLANG-1M-VQA} test set. Compared to the performance before fine-tuning, the LLaVA1.5-7B \citep{liu2024visual} model exhibited a significant improvement in overall accuracy, with an increase of 36.75\% after fine-tuning. Similar performance improvements were observed for the QWEN2-VL-7B \citep{wang2024qwen2}, QWEN2.5-VL-7B \citep{Qwen2.5-VL}, and LLaVA1.5-13B \citep{liu2024visual} models, with overall accuracy enhancements of 40.22\%, 34.73\%, and 40.13\%, respectively. These results demonstrate that our dataset substantially enhances the performance of VLMs on SAR VQA task. 

\begin{table*}[ht]
\renewcommand{\arraystretch}{1.35}
\centering
\caption{\label{tab:table41}{\bf SAR Image Captioning Results on \textit{SARLANG-1M-Cap} Benchmark.} 'Pre' represents preprocessing strategies, while '\checkmark/\ding{55}' denote the model evaluation with/without SAR image preprocessing.}
\scalebox{0.9}{
\begin{tabular}{|l|l|l|l|c|c|c|c|c|c|}
\hline
\textbf{Caption} & \textbf{Model}& \textbf{Param} & \textbf{Pre} & \textbf{BLEU\_1} & \textbf{BLEU\_2} & \textbf{BLEU\_3} & \textbf{BLEU\_4} & \textbf{ROUGE\_L} & \textbf{CIDEr} \\ \hline
\multirow{2}{*}{Complex} 
& QWEN2.5-VL \citep{Qwen2.5-VL} & 3B & \ding{55} & 26.09 & 14.92 & 9.87  & 6.11  & 25.31 & 12.40 \\ 
& QWEN2.5-VL \citep{Qwen2.5-VL} & 3B & \checkmark & \textbf{26.95} & \textbf{16.44} & \textbf{11.10} & \textbf{6.65}  & \textbf{26.45} & \textbf{14.78} \\ 
\hline
\multirow{2}{*}{Concise} 
& QWEN2.5-VL \citep{Qwen2.5-VL} & 3B & \ding{55} &  25.18 & 15.57 & 10.84 & 7.35  & 23.73 & 5.79\\
& QWEN2.5-VL \citep{Qwen2.5-VL} & 3B & \checkmark & \textbf{30.57} & \textbf{21.54} & \textbf{15.97} & \textbf{10.97}  & \textbf{29.36} & \textbf{6.25} \\ 

\hline
\end{tabular}}
\end{table*}
\begin{table}[ht]
\renewcommand{\arraystretch}{1.35}
\centering
\caption{\label{tab:table81}{\bf SAR VQA Results on \textit{SARLANG-1M-VQA} Benchmark.} 'Pre' represents preprocessing strategies, while '\checkmark/\ding{55}' denote the model evaluation with/without SAR image preprocessing. The accuracy metric denotes the ratio of correctly answered questions.}
\scalebox{0.93}{
\begin{tabular}{|l|l|l|c|}
\hline
\textbf{Model} & \textbf{Param} & \textbf{Pre}  & \textbf{Accuracy} \\ \hline
QWEN2.5-VL \citep{Qwen2.5-VL} & 3B & \ding{55} & 29.51  \\ 
QWEN2.5-VL \citep{Qwen2.5-VL} & 3B & \checkmark & \textbf{32.32}  \\ 
\hline
\end{tabular}}
\end{table}

\par To further compare the SAR image interpretation ability of VLMs with humans, we meticulously selected 30 diversified question-answering pairs from the \textit{SARLANG-1M-VQA} dataset, serving as the \textit{validation} set. Three SAR experts and three normal persons were invited to answer these questions. Human scores, including ordinary people’s scores and expert’s scores, are computed as the average ratio of correctly answered questions for each group. Table \ref{tab:table82} presents the VQA performance of various VLMs and human scores. The results indicate that, after fine-tuning on the \textit{SARLANG-1M-VQA} dataset, QWEN2.5-VL-7B \citep{Qwen2.5-VL} model outperforms both ordinary individuals and SAR experts. This further demonstrates that mainstream VLMs can overcome non-expert barriers in SAR image interpretation and achieve a level of understanding comparable to human experts.

\par We also present a visualization of selected SAR VQA results from the \textit{SARLANG-1M-VQA} benchmark in Figure \ref{fig:fig38}. The experimental results demonstrate significant performance improvements for the LLaVA1.5-7B \citep{liu2024visual} model and QWEN2.5-VL-7B \citep{Qwen2.5-VL} model across applications 2-7 after fine-tuning. A notable example is observed in application 4, where the pre-trained LLaVA1.5-7B \citep{liu2024visual} model fails to accurately quantify targets in SAR images. However, after fine-tuning on our dataset's training set, the LLaVA1.5-7B \citep{liu2024visual} model achieves precise quantification of the 'terminal' target class. Besides, application 7 addresses a fundamental remote sensing task: land cover classification. When processing SAR images of agricultural areas dominated by crops with sparse distributions of trees and grasslands, the pre-trained LLaVA1.5-7B \citep{liu2024visual} model and QWEN2.5-VL-7B \citep{Qwen2.5-VL} model initially classify the entire region as 'open field' and 'a small white dot', respectively. Following fine-tuning on our \textit{SARLANG-1M-VQA} dataset, the LLaVA1.5-7B \citep{liu2024visual} model and the QWEN2.5-VL-7B \citep{Qwen2.5-VL} model demonstrate enhanced capability, accurately identifying the primary land use patterns. These experimental results demonstrate that our \textit{SARLANG-1M-VQA} dataset significantly enhances the performance of existing VLMs on SAR image understanding tasks.


\subsection{Necessity of Preprocessing}

\par In this section, we analyze the impacts of SAR preprocessing operations, specifically denoising and contrast stretching, on the performance of VLMs in SAR image captioning and VQA tasks. The QWEN2.5-VL-3B \citep{Qwen2.5-VL} model, without fine-tuning on our \textit{SARLANG-1M} dataset, is chosen for evaluation.

\subsubsection{SAR Image Captioning on \textit{SARLANG-1M-Cap} Benchmark}

\par To demonstrate the effectiveness of preprocessing operations on the SAR image captioning task, we conduct two comparative experiments to quantify the contribution of the SAR image preprocessing strategy. We selected 1,170 complex captions and 2,571 concise captions from the test set of the \textit{SARLANG-1M-Cap} benchmark, serving as ground truth for SAR images from the OpenEarthMap-SAR \citep{xia2025openearthmap} dataset. As shown in Table \ref{tab:table41}, preprocessing the SAR images significantly enhances the performance of the QWEN2.5-VL-3B \citep{Qwen2.5-VL} model, particularly for concise captions. Specifically, the model achieves improved scores of 5.39 for BLEU\_1, 5.63 for ROUGE\_L, and 0.46 for CIDEr. These results confirm the effectiveness of our SAR image preprocessing operations on the VLM's performance in the SAR image captioning task.


\subsubsection{SAR VQA on \textit{SARLANG-1M-VQA} benchmark}

To further confirm the effectiveness of preprocessing operations in the SAR VQA task, 9,220 VQA labels are selected from the test set of the \textit{SARLANG-1M-VQA} benchmark, serving as ground truth for SAR images from the OpenEarthMap-SAR \citep{xia2025openearthmap} dataset. As shown in Table \ref{tab:table81}, our SAR image preprocessing operations result in an overall accuracy improvement of 2.81 for the QWEN2.5-VL-3B \citep{Qwen2.5-VL} model. The performance improvements are achieved by the denoising operation, which removes elements like speckle noise that hinder VLMs comprehension of SAR images. Additionally, linear stretching enhances SAR image contrast, enabling VLMs to focus more effectively on salient objects and areas. These results further validate the necessity of preprocessing operations prior to evaluating VLMs on SAR images.

\section{Conclusion}
\label{sec:conclusion}
In this paper, we introduced \textit{SARLANG-1M}, a large-scale dataset designed to advance multimodal understanding of SAR images. \textit{SARLANG-1M} comprises approximately 1 million high-quality SAR image-text pairs collected from over 59 cities worldwide, offering multi-scale resolutions ranging from 0.1 to 25 meters, fine-grained semantic descriptions (including both concise and complex captions), and diverse remote sensing categories (1,696 object categories and 16 land cover categories). The dataset features two benchmarks, \textit{SARLANG-1M-Cap} and \textit{SARLANG-1M-VQA}, which support seven key remote sensing applications, including image description, object identification, object classification, instance counting, region referring, object positioning, and others. Extensive experiments demonstrate that fine-tuning existing VLMs on \textit{SARLANG-1M} significantly enhances their performance in SAR image understanding tasks, achieving interpretative abilities comparable to human experts. Further, our SAR image preprocessing strategies effectively enhance the VLMs' ability in SAR image interpretation.

\section*{Acknowledgements}
\par This work was supported in part by the Council for Science, Technology and Innovation (CSTI), the Cross-ministerial Strategic Innovation Promotion Program (SIP), Development of a Resilient Smart Network System against Natural Disasters (Funding agency: NIED), the JSPS, KAKENHI under Grant Number 25K03145, 22H03609, 24KJ06\-52, JST, FOREST under Grant Number JPMJFR206S, and Next Generation AI Research Center of The University of Tokyo.










\appendix

\section{\textit{Overall Image Similarity}} \label{app:ImageSim}

\par Given a image set $[{f}_1,{f}_2,...,{f}_N]$ with $N$ images, the \textit{overall image similarity} ${S}_{Image}=(\mu, \sigma)$ is computed as follows:
\begin{align}
V_i = \text{Feature\_Extractor}[{f}_i],
\end{align}
\begin{align}
V_t = \text{Feature\_Extractor}[{f}_t],
\end{align}
where $V_i$ and $V_t$ correspond to the extracted image features of ${f}_i$ and ${f}_t$.The VGG16 \citep{simonyan2014very} model is designated as {Feature\_Extractor}.

Following image feature extraction, the cosine similarity between each image pair $({f}_i,{f}_t)$ is then calculated based on the extracted features$({V}_i,{V}_t)$:
\begin{align}
{S}_{i,t} = \frac{V_i \cdot V_t}{\|V_i\| \|V_t\|},
\end{align}
where $\cdot$ denotes the dot product, and $\|\cdot\|$ represents the Euclidean norm of the feature vectors. 

The cosine similarities ${S}_{i,t}$ computed for all image pairs form an image similarity matrix $S$. To prevent redundant calculations, only the lower triangle of $S$ is retained, creating an image similarity set $s$:
\begin{align}
{s} = \{ S_{it} \mid i > t \}.
\end{align}

Then we compute the mean $\mu$ and variance $\sigma^2$ of the similarity set $s$ to represent the \textit{overall image similarity} $S_{Image}$ of the dataset:
\begin{align}
\mu = \frac{1}{M} \sum_{k=1}^{M} s_k,
\end{align}
\begin{align}
\sigma^2 = \frac{1}{M} \sum_{k=1}^{M} (s_k - \mu)^2,
\end{align}
\begin{align}
S_{Image}=(\mu, \sigma),
\end{align}
where $M$ indicates the number of image pairs.

\section{\textit{Overall Text Similarity}}\label{app:TextSim}
\par Given a text set $[{h}_1,{h}_2,...,{h}_{P}]$ with $P$ text, the \textit{overall text similarity} ${S}_{Text}=(\mu_1, \sigma_1)$ is defined as follows:
\begin{align}
E_j = \text{Feature\_Extractor}[{h}_j],
\end{align}
\begin{align}
E_k = \text{Feature\_Extractor}[{h}_k],
\end{align}
where $E_j$ and $E_k$ are extracted features of the text ${h}_j$ and ${h}_k$, respectively. The Sentence-BERT \citep{reimers2019sentence} model is denoted as {Feature\_Extractor}. The cosine similarity between the text embeddings text pairs $({h}_j,{h}_k)$ is calculated as
\begin{align}
{L}_{j,k} = \frac{E_j \cdot E_k}{\|E_j\| \|E_k\|}.
\end{align}
Then the lower triangle of the text similarity matrix $L$ constitutes a text similarity set $l$.
\begin{align}
{l} = \{ L_{jk} \mid j > k \}.
\end{align}

The mean $\mu_1$ and variance $\sigma_1^2$ are calculated from the similarity set $l$ of text pairs:
\begin{align}
\mu_1 = \frac{1}{M_1} \sum_{k=1}^{M_1} s1_k,
\end{align}
\begin{align}
\sigma_1^2 = \frac{1}{M_1} \sum_{k=1}^{M_1} (l_k - \mu_1)^2,
\end{align}
\begin{align}
{S}_{Text}=(\mu_1, \sigma_1),
\end{align}
where $M_1$ represents the number of text pairs.



\printcredits

\bibliographystyle{cas-model2-names}

\bibliography{main}





\end{document}